\documentclass[10pt,twocolumn,letterpaper]{article}

\usepackage{cvpr}              

\usepackage{multicol}
\usepackage{multirow}
\usepackage{color, colortbl}
\usepackage{bbding}
\usepackage{booktabs}
\usepackage{bbding} 
\usepackage{pifont}
\usepackage{makecell} 

\usepackage{xcolor}
\definecolor{checked}{rgb}{0.7, 0.7, 0.2} 

\definecolor{cvprblue}{rgb}{0.21,0.49,0.74}
\usepackage[pagebackref,breaklinks,colorlinks,allcolors=cvprblue]{hyperref}

\def\paperID{2524} 
\def\confName{CVPR}
\def\confYear{2025}

\title{DexGrasp Anything: Towards Universal Robotic Dexterous Grasping \\ with Physics Awareness}

\author{
Yiming Zhong$^{1,*}$,
Qi Jiang$^{1,*}$,
Jingyi Yu$^{1}$, 
Yuexin Ma$^{1,\dagger}$
\\ 
$^{1}$ ShanghaiTech University\\
\{zhongym2024, jiangqi2022, yujingyi, mayuexin\}@shanghaitech.edu.cn
}

\begin{document}

\makeatletter
\let\@oldmaketitle\@maketitle
\renewcommand{\@maketitle}{
   \@oldmaketitle
 \begin{center}
   \vspace{-5ex}
      \includegraphics[width=1.0\linewidth]{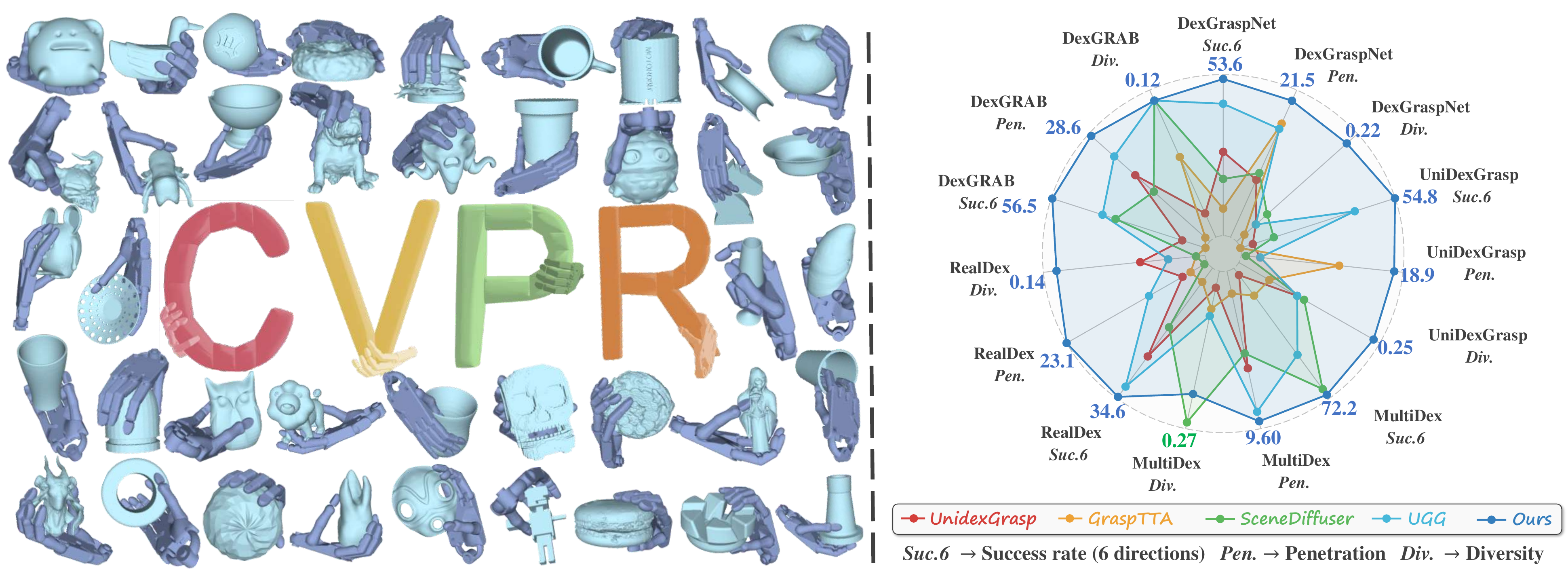}
 \end{center}
 \vspace{-2ex}
  \refstepcounter{figure}\normalfont Figure~\thefigure. 
  We present \textbf{DexGrasp Anything}, consistently surpassing previous dexterous grasping generation methods across five benchmarks. Visualization of our method's results are shown on the left.
  \label{fig:teaser}
  \newline
  }


\maketitle
\footnotetext[1]{$^{*}$ Equal contribution.}
\footnotetext[2]{$\dagger$ Corresponding author. This work was supported by NSFC (No.62206173), Shanghai Frontiers Science Center of Human-centered Artificial Intelligence (ShangHAI), MoE Key Laboratory of Intelligent Perception and Human-Machine Collaboration (KLIP-HuMaCo).}

\begin{abstract}
   A dexterous hand capable of grasping any object is essential for the development of general-purpose embodied intelligent robots. However, due to the high degree of freedom in dexterous hands and the vast diversity of objects, generating high-quality, usable grasping poses in a robust manner is a significant challenge. In this paper, we introduce DexGrasp Anything, a method that effectively integrates physical constraints into both the training and sampling phases of a diffusion-based generative model, achieving state-of-the-art performance across nearly all open datasets. Additionally, we present a new dexterous grasping dataset containing over 3.4 million diverse grasping poses for more than 15k different objects, demonstrating its potential to advance universal dexterous grasping. Code and dataset are available at \url{https://github.com/4DVLab/DexGrasp-Anything}
\end{abstract}
\vspace{-4ex}

\section{Introduction}
\label{sec:intro}
Dexterous grasping, serving as a foundational capability for robotic manipulation tasks, has attracted significant attention. Five-fingered dexterous hands, which closely resemble the structure of the human hand, offer significantly greater flexibility, manipulation precision, and versatility compared to simpler grippers (e.g. parallel jaw, vacuum gripper). As robots are being deployed in human environments, dexterous hands have become increasingly vital due to their ability to interact with a wide range of objects as humans and use tools designed for humans. 
Therefore, a precise, robust, and versatile dexterous grasping method lies at the heart of interactions in embodied intelligence.

Earlier approaches~\cite{ap1,ap2,ap3,ap4,ap5,ap6} on dexterous grasping have primarily relied on analytical methods, where grasping poses were optimized to meet specific physical constraints. These methods, however, face significant challenges due to the large search space and the complexity of optimizing for high degrees of freedom in dexterous hands, leading to low success rates. 
In contrast, data-driven methods~\cite{ddg,DGTR,xu2023unidexgrasp,grasptta,lu2023ugg,scenediffuser,weng2024dexdiffuser,GYS,zhang2024dexgrasp-diffusion} leverage large-scale datasets to learn useful priors, narrowing the search space and providing strong guidance for search initialization. 
Regression-based methods~\cite{ddg,DGTR} that directly predict grasp parameters from object inputs often suffer from mode collapse and averaging, resulting in a limited diversity of generated grasp poses. Recently, generative methods~\cite{grasptta,xu2023unidexgrasp} have gained significant attention for their ability to enhance the diversity of generated grasp poses. Among these, diffusion models~\cite{ddpm,song2020sde,karras2022elucidating} have demonstrated a strong capability to capture the complexities of dexterous grasping and generate diverse grasp poses by iteratively transforming a simple distribution (e.g., Gaussian) into a complex, high-dimensional one~\cite{weng2024dexdiffuser,scenediffuser,lu2023ugg,zhang2024dexgrasp-diffusion}. However, despite these advantages, current diffusion-based approaches~\cite{GYS,DGTR} often generate suboptimal grasp poses, resulting in hand-object penetration or insufficient contact with unsatisfactory success rates. These issues arise from the lack of constraints that enforce physical rules.


In this work, we propose a novel dexterous grasping generation method, namely \textbf{DexGrasp Anything}, that integrates three carefully designed physical constraint objectives into the diffusion model during both training and sampling phases. DexGrasp Anything exhibits superior robustness and strong generalization capabilities. Specifically, we introduce the \textit{surface pulling force} to ensure grasp feasibility by pulling the hand's inner surface toward the object’s surface while avoiding interference with parts that are already sufficiently distant. We also introduce the \textit{external-penetration repulsion force} to maintain geometric accuracy of interaction by effectively preventing significant collisions between the object and the dexterous hand, and the \textit{self-penetration repulsion force} to preserve the hand's geometry by enforcing a minimum distance between finger joints and applying repulsion when they get too close. Through our \textbf{physics-aware training scheme} and \textbf{physics-guided sampler}, these physical constraints enable our diffusion-based generator to produce practical and robust dexterous grasping poses across a wide range of objects. Through extensive experiments, we show our method achieves state-of-the-art performance on almost all open datasets, as Figure~\ref{fig:teaser} shows.


To further improve the universality of diffusion-based generative method, massive amounts of high-quality training data is necessary. While many efforts have been made to build grasping datasets, they suffer from narrow data distribution\cite{ddg,lundell2021ddgc,graspit-data2}, limited object categories~\cite{ddg,wang2023dexgraspnet}, and scalability issues~\cite{liu2024realdex}.
In light of this, we dedicate substantial efforts to further enhancing the scale, diversity, and quality of the dexterous grasping datasets. We start by gathering available dexterous grasping data from multiple sources~\cite{gendexgrasp-multidex,xu2023unidexgrasp,wang2023dexgraspnet,liu2024realdex,taheri2020grab}, including simulated data, real-captured data, and human hand grasping data, ensuring a diverse and comprehensive data distribution.
We further scale up the dataset with a `model-in-the-loop' strategy by using our grasping method and filtering method to continue generating high-quality data, inspired by the approach used in SAM~\cite{kirillov2023sam}. These efforts culminates in a very large-scale dexterous grasping dataset,  \textbf{DexGrasp Anything (DGA) Dataset} with over 3.4 million grasping poses on more than 15k objects. Experimental results demonstrate that this new dataset provides substantial benefits to grasping methods within the community.

The main contribution of this work are as follows:
\begin{itemize}
    \item  We propose a physics-aware diffusion generator for dexterous grasping pose generation, which effectively integrates three key physical constraints into both the training and sampling phases of the diffusion model. 
    \item Our method achieves state-of-the-art performance on five dexterous grasping datasets.
    \item We present a new high-quality dexterous grasping dataset, the largest and most diverse to date, significantly improving the generalization capability of existing methods.
\end{itemize}

\section{Related Work}
\label{sec:related}

\subsection{Dexterous Grasp Generation}
Dexterous Grasping serves as a fundamental component for various complex, human-like manipulation tasks, making it a long-standing area of research in robotics. Early works mainly use manually derived analytical methods~\cite{ap1,ap2,ap3,ap4,ap5,ap6} that based on certain physical constraints. These methods are hindered by extremely large search space and complex optimization process, leading to low success rate.

Recently, data-driven approaches have emerged as a promising direction for dexterous grasping. 
Regression-based methods, such as \cite{ddg, DGTR}, often generate grasping poses with limited diversity, as they rely on direct predictions from input data and may fail to explore the full range of possible grasping configurations. In contrast, generative methods~\cite{grasptta, xu2023unidexgrasp} explicitly model the conditional probability distribution of dexterous hand poses given the target object, theoretically generating diverse poses. Diffusion model-based methods~\cite{lu2023ugg,zhang2024dexgrasp-diffusion,scenediffuser} , in particular, stand out as a promising direction for more universal and robust robotic dexterous grasping for its exceptional capabilities in modeling various complex data distribution~\cite{Liu2023AudioLDMTG,Huang2023MakeAnAudioTG,Liu2024SoraAR,zhang2024clay,Ren2023XCubeL,Xu2022GeoDiffAG} and generating diverse and highly realistic samples~\cite{peebles2023dit, Ramesh2022HierarchicalTI,saharia2022imagen,latent-diffusion}.
However, existing diffusion model-based approaches are observed~\cite{GYS,DGTR} to yield sub-optimal grasping poses due to the absence of physical constraints during the training and sampling process.
In this work, we delve into incorporating physical constraints into diffusion models to generate robust dexterous grasping poses for dexterous hands.

\begin{figure*}[t]
  \centering
   \includegraphics[width=1.0\linewidth]{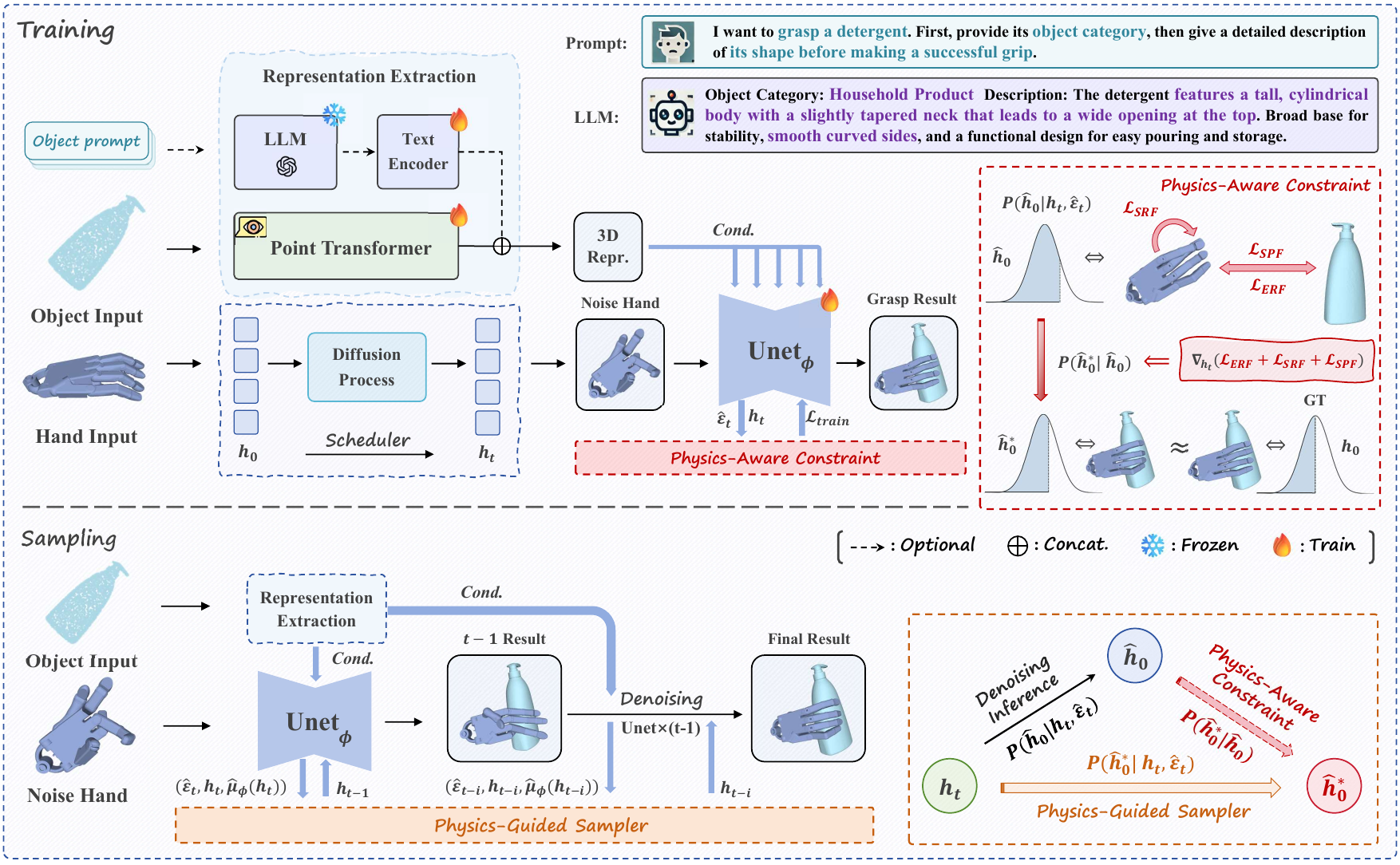}

   \caption{Overview of DexGrasp Anything. During training, object information is processed to extract combined semantic and spatial representations as conditioning inputs. At each noise training step, a clean-estimation of the noisy hand pose \( \hat{h}_0 \) is derived from the predicted noise \( \hat{\varepsilon}_t \), with physics constraints guiding the noise distribution toward a cleaner, grasp-suitable distribution. During sampling, the Physics-Guided Sampler obtains the current observation \( \hat{h}_0 \) at each denoising step and performs posterior sampling based on this observation. Physical constraints gradually guide the distribution toward a physically feasible grasp configuration \( \hat{h}_0^* \), enabling effective grasping of diverse objects.
}
   \label{fig:pipline}
   \vspace{-2ex}
\end{figure*}

\subsection{Dexterous Grasping Dataset}
Collecting 3D dexterous grasping poses is notoriously expensive due to the complexity of hand structures. Most existing datasets are collected in simulators like GraspIt!~\cite{miller2004graspit} and IssacGym~\cite{makoviychuk2021isaac} through searching in the eigengrasp space~\cite{ddg,lundell2021ddgc,graspit-data2} or optimization-based methods~\cite{gendexgrasp-multidex,ddg,xu2023unidexgrasp,wang2023dexgraspnet} in parameter space. However, the search-based data often follow a narrow distribution due to the low-dimensional eigengrasp space, while the recent optimization-based data still suffer from relatively success rate and contained limit number of object categories.
Some real-world datasets have been collected using teleoperation systems controlled by human operators. While these datasets capture human-like grasping poses, the data collection process is prohibitively expensive and difficult to scale up. Recent advances~\cite{qin2023anyteleop,shaw2023videodex, sivakumar2022robotic} have explored  retargeting human hand poses to dexterous robotic hands, presenting a promising avenue for leveraging human hand data in robot hand training.
Despite these advancements, existing datasets still face limitations in diversity, scalability, and quality. To address these challenges, we introduce the largest and most diverse dexterous grasping dataset to date and demonstrate that our dataset significantly enhances both the quality and diversity of generated dexterous grasping poses, providing substantial value to existing data-driven grasping generators.\\


\section{Method}
\label{sec:methods}
We present \textbf{DexGrasp Anything}, an effective approach that enhances diffusion-based generators for dexterous grasping by integrating meticulously designed physical constraints. Figure \ref{fig:pipline} demonstrates an overview of our method.
The following sections detail our problem formulation (Sec.~\ref{sec:problem-definition}), physics-aware constraints (Sec.~\ref{sec:physical-constraints}), and the way to integrate constraints to the diffusion model's training (Sec.~\ref{sec:training}) and sampling processes (Sec.~\ref{sec:sampler}), and the LLM-enhanced object representation extraction module (Sec.~\ref{sec:representation}).

\subsection{Problem Definition}\label{sec:problem-definition}
Our goal is to generate high-quality grasping poses capable of securely holding a given object.  Specifically, given a 3D object observation $O$, we aim to sample a diverse set of dexterous grasping poses $\mathbf{h} = \{h_i\}_{i=1}^{n}$ from a conditional distribution $P(h|O)$, where the dexterous pose $h = (\theta,\mathrm{R}, \mathrm{t}) \in \mathbb{R}^{33}$  consists of the dexterous hand articulation $\theta \in \mathbb{R}^{24}$(for ShadowHand), the global rotation $R \in \mathrm{SO(3)}$ and the global translation vector $t \in \mathbb{R}^3$. 

The conditional distribution $P(h|O)$ is modeled using a diffusion model $\epsilon_{\phi}(h_t,O,t)$, which iteratively transforms an isotropic Gaussian distribution $\mathcal{N}(0, \mathrm{I})$ into the desired data distribution:
\begin{equation}
\begin{split}
P(h_0|O) &= P(h_T)\prod_{t=1}^{T}P(h_{t-1}|h_t, O), \\
P(h_{t-1}|h_t, O) &= \mathcal{N}(h_{t-1} ; \mu_{\phi}(h_t,O,t), \Sigma_{\phi}(h_t,O,t)).
\end{split}
\end{equation}

\subsection{Physical Constraints}\label{sec:physical-constraints}
Diffusion model-based methods often fail to reach optimal performance in the absence of appropriate physical constraints. To address this, we present three tailored physical constraint objectives for our DexGrasp Anything generator, enabling the production of universal and robust dexterous grasping poses for a wide range of objects.

\noindent\textbf{Surface Pulling Force}~\cite{xu2023unidexgrasp} is crucial for ensuring grasp feasibility. 
It enforces proximity between the inner surface of the robotic phalanges (represented by sampled point clouds) and the object's surface. This guidance signal applies a pulling force only to points that are closer than a specified threshold, ensuring that the points on the inner surface of the fingers are pulled towards the object's surface when they are near, but does not affect points that are already at a sufficient distance. 
Compute the squared Euclidean distance for each inner surface point $p_{\text{dis}}^{(i)}$ to its nearest neighbor on the object surface:
$d_i = \min_{j} \| p_{\text{dis}}^{(i)} - p_{\text{obj}}^{(j)} \|^2$
We compute the surface pulling force as:
\begin{equation}
    L_{\text{SPF}} = \frac{\sum_{i \in S} \sqrt{d_i}}{|S| + \eta},
\end{equation}
where $S =\{i| d_i < d_{\text{Threshold}} \}$ represents the set of points within the threshold distance and $\eta$ is a numerical stability constant.

\noindent\textbf{External-penetration Repulsion Force}~\cite{scenediffuser} retains the spatial accuracy of hand-object interactions. It minimizes the undesired intersection between the hand and object point clouds by leveraging the signed distances.
Given an object point cloud $P_{obj}$ with surface normals $N_{obj}$ , and a hand point cloud $P_{hand}$, we first compute the nearest neighbor distance between each point in $P_{hand}$ and $P_{obj}$ : $d_{i} = \min_{j} \| P_{hand}^{(i)} - P_{obj}^{(j)} \|$. 
We then calculate the signed distance between the hand and object points using the object normals:
\begin{equation}
    s_{i} = \text{sign}((P_{obj}^{(j)} - P_{hand}^{(i)}) \cdot N_{obj}^{(j)}).
\end{equation}
Finally, the external-penetration repulsion force is defined as the maximum signed distance across all points in the batch, averaged over the batch size $B$:
\begin{equation}
    L_{\text{ERF}} = \frac{1}{B}\sum_{i=1}^{B} max_i(s_i,d_i).
\end{equation}

\noindent\textbf{Self-Penetration Repulsion Force}~\cite{xu2023unidexgrasp} upholds the hand structural geometry. It addresses the issue of hand points intersecting with each other by enforcing a minimum distance between them. This ensures that the hand maintains a realistic, physically plausible shape without finger collisions.
Given a set of hand points $P_{\text{hand}}$, we calculate the pairwise Euclidean distances between all points:
$d_{ij} = \| P_{\text{hand}}^{(i)} - P_{\text{hand}}^{(j)} \|$,
A repulsion force is applied when $d_{ij}$ is smaller than a threshold $d_{\text{Threshold}}$, and the self-penetration repulsion force is defined as:
\begin{equation}
    L_{\text{SRF}} = \frac{1}{B} \sum_{i,j} \max(0, d_{\text{Threshold}} - d_{ij}).
\end{equation}



\subsection{Physics-Aware Training}\label{sec:training}
Diffusion Models are typically trained with a simple mean-squared error (MSE) objectives:
\begin{equation}
    L_{\mathrm{simple}} = \mathbb{E}_{t,x_0,\epsilon}[ \| \epsilon - \epsilon_{\phi}(h_t,t)\|^2],
\end{equation}
where $h_t$ is a corrupted version of the original data $h_0$. The data corruption follows a fixed noise schedule $\beta_t$:
\begin{equation}
    \begin{split}
        q(h_t | h_{t-1}) &= \mathcal N (h_t;\sqrt{1-\beta_t}h_{t-1}, \beta_t\textbf{I}), \\
        h_t &= \sqrt{1-\beta_t} h_{t-1} + \sqrt{\beta_t} \epsilon_t.
    \end{split}
\end{equation}
With $\alpha_t=1-\beta_t$ and $\bar{\alpha}_t=\Pi_{i=1}^t \alpha_i$, the corruption process can be directly conditioned on $h_0$:
\begin{equation}\label{equ:forward}
\begin{split}
q(h_t | h_{0}) &= \mathcal N (h_t;\sqrt{\bar{\alpha}_t}h_{0}, (1-\bar{\alpha}_t)\textbf{I}), \\
h_t &= \sqrt{\bar{\alpha}_t} h_{0} + \sqrt{1-\bar{\alpha}_t} \bar{\epsilon_t}.
\end{split}
\end{equation}

Since the training objective $L_{\text{simple}}$ is essentially a reweighted variational lower bound on negative log likelihood, it does not incorporate any explicit supervision regarding physical constraints. As a result, diffusion models often perform sub-optimally when directly generating dexterous grasping parameters, as observed in previous works~\cite{GYS,DGTR}. To facilitate our diffusion generator in capturing physics priors during training, we introduce the \textbf{Physics-Aware Training} paradigm, which incorporates the tailored physical constraints outlined in Sec.~\ref{sec:physical-constraints}.

The training objective involving only corrupted data $h_t$ is not well-suited for incorporating physical constraints.
Using the diffusion process defined in Eq.~\ref{equ:forward}, the estimated sample:
$\hat{h_0}(h_t) = \frac{1}{\sqrt{\bar{\alpha}_t}}(h_t - \sqrt{1-\bar{\alpha}_t} \epsilon_{\phi}(h_t,O,t))$ 
could serve as a good proxy for introducing physical constraint into the training process of a diffusion model.

We define the physically-aware training objective $L_{\text{PADG}}$ as a linear combination of the standard mean-squared objective and multiple physical-constraint objectives:
\begin{equation}
    L_{\text{PADG}} = L_{\text{simple}}(h_t) + \sum_{i=1}^m \alpha_i L_{\text{PA}_i}(\hat{h_0}(h_t), \epsilon_{\phi}),
\end{equation}
where $L_{\text{PA}_i}(\hat{h_0}(h_t), \epsilon_{\phi})$ is the $i^{th}$ physical constraint and $\alpha_i$ is the corresponding weighting coefficient. The gradient is propagated to $h_t$ through the estimated clean sample $h_0$ as follow:
\begin{equation}
    \frac{\partial L_{\text{PADG}}}{\partial h_t} = \frac{\partial L_{\text{simple}}}{\partial h_t} + \sum_{i=1}^m \alpha_i \frac{\partial L_{\text{PA}_i}}{\partial \hat{h_0}} \cdot \frac{\partial \hat{h_0}}{\partial h_t}.
\end{equation}


\subsection{Physics-Guided Sampling}\label{sec:sampler}
Leveraging the learned physics priors, the well-trained diffusion generator is capable of producing physically plausible dexterous grasping poses for a given object. The physical constraints can be further enhanced during the sampling process by employing advanced sampling techniques~\cite{Dhariwal2021Diffusionbeatgan,Ho2022ClassifierFreeDG,Yang2024diffusion-gaussian-constraint,Chung2022DiffusionPS}. 

Classifier guidance~\cite{Dhariwal2021Diffusionbeatgan} has explored the use of a time-dependent classifier $\mathcal F_t$ to steer the diffusion model towards specific conditional distributions. The guidance can be approximated as an offset in the posterior mean:
\begin{equation}
    \widetilde{\mu}_{\phi}(x_t,t) \leftarrow \mu_{\phi}(x_t,t) + s\Sigma_{\theta, t}\nabla_{x_t} log(\mathcal F_t(y,x_t)),
\end{equation}
where $s$ is the guidance strength. By estimating $\hat{x_0}$ based on $x_t$, the guidance signal can be extended from a time-dependent classifier $\mathcal F_t(y,x_t)$ to arbitrary objective functions $L(x_0)$ on clean samples. It can be achieved by mapping the objective to a density-like function: 
\begin{equation}
    \mathcal{F}(x_0) = \mathcal{Z}e^{-L(x_0)},
\end{equation}
where $\mathcal Z$ is a normalizing constant.
We define the Physics-Guided Sampler as follows:
\begin{equation}
\begin{split}
    \widetilde{\mu}_{\phi}(h_t,O,t) & \leftarrow \mu_{\phi}(h_t,O,t) \\
    & + s\Sigma_{\phi,t}  \nabla_{h_t} \sum_{i=1}^m \alpha_i L_{PA_i}(\hat{h_0}(h_t),\epsilon_t).
\end{split}
\end{equation}
To alleviate the estimation bias on $\hat{h_0}$, we apply the Spherical Gaussian Constraint~\cite{Yang2024diffusion-gaussian-constraint} with a weighted gradient direction in practice. This is expressed as:
\begin{equation}
    \widetilde{\mu}_{\phi}(h_t,O,t) \leftarrow \mu_{\phi}(h_t,O,t) + r\frac{d_m}{\| d_m \|},
\end{equation}
where $d_\text{m} = d_{\text{sample}} + g_r(d^{*} - d_{\text{sample}})$, $d_{\text{sample}} = \Sigma_{\phi,t}\epsilon$ and:
\begin{equation}
    \begin{split}
        d^* &= -\sqrt{n}\Sigma_{\phi,t}\frac{\nabla_{h_t} \sum_{i=1}^m \alpha_i L_{\text{PA}_i}(\hat{h_0}(h_t),\epsilon_t)}{\| \nabla_{h_t} \sum_{i=1}^m \alpha_i L_{\text{PA}_i}(\hat{h_0}(h_t),\epsilon_t) \|}.
    \end{split}
\end{equation}

Incorporating physics constraints during training helps guide the noise distribution toward a cleaner, grasp-suitable form. However, due to sparse supervision in the training phase, we leverage the reverse process during sampling to obtain \( h_t \) and \( \hat{h}_0 \) at each step, applying posterior sampling to iteratively refine the grasp configuration. This iterative refinement allows the Physics-Guided Sampler to progressively adjust \( \hat{h}_0 \) under physics constraints, ultimately steering the distribution toward a physically feasible grasp configuration. Through this distributional framework, our model effectively generalizes to diverse objects, demonstrating robustness and adaptability in grasping tasks.
\subsection{LLM-enhanced Representation Extraction}\label{sec:representation}
To accomplish robust dexterous grasp generation for a targeted object, we boost the traditional object representation by complementing the geometry object feature with semantic prior from powerful LLMs.
We employ a Point Transformer\cite{zhao2021point-transformer} to encode object point clouds, producing an 
$N\times C$ feature vector, where N represents the number of groups defined by the Point Transformer. To enrich these features with abundant semantic prior from LLM, we utilize the prompt: ``\textit{I want to grasp a [object label]. First, provide its object category, then give a detailed description of its shape before making a successful grip.}"  We then parse the response and encode each sentence using a pre-trained \textit{BERT-large-uncased} model. We extract the [CLS] token from each sentence and apply max-pooling on them. This results in an $1\times C$ semantic feature vector that includes the rich prior knowledge from the LLM. The concatenated $(N+1)\times C$ feature matrix is subsequently integrated into the diffusion backbone through a cross-attention mechanism, enhancing the model's capacity to generate precise and contextually relevant grasping poses.


\section{Dataset}
\label{sec:Dataset}

\begin{table*}[htbp]
\centering
\caption{Comparison of dexterous grasp datasets. Our dataset achieves the largest scale to date.}
\vspace{-1ex}
\resizebox{\textwidth}{!}{  
\begin{tabular}{@{}lccccccc@{}}  
\toprule
Dataset & Publication Year & Hand Type & Sim./Real & \#Grasps & \#Objects & \#Grasps per object & Construction method  \\ 
\midrule
GRAB~\cite{taheri2020grab} & ECCV 2020 & MANO  & Real & 1.64M & 51 & \textgreater10K & Capture  \\  
DDGdata~\cite{ddg} & RSS 2020 & ShadowHand & Sim. & 6.9k & 565 & - & GraspIt! \\ 
MultiDex~\cite{gendexgrasp-multidex} & ICRA 2023 & ShadowHand & Sim. & 16K & 58 & 300 & Optimization  \\ 
DexGraspNet~\cite{wang2023dexgraspnet} & ICRA 2023 & ShadowHand & Sim. & 1.32M & 5,355  & \textgreater200 & Optimization  \\ 
UniDexGrasp~\cite{xu2023unidexgrasp} & CVPR 2023 & ShadowHand & Sim. & 1.12M & 5,519  & \textgreater150 & Optimization \\ 
GraspXL~\cite{zhang2024graspxl} & ECCV 2024 & Diverse Hand & Sim & - & \textbf{500K} & - & GraspXL \\
RealDex~\cite{liu2024realdex} & IJCAI 2024 & ShadowHand & Real & 59K & 52 & \textgreater200 & Human Annotation \\ 
\midrule
Our dataset & CVPR 2025 & ShadowHand & \textbf{Real+Sim.} & \textbf{3.40M} & 15,698 & \textgreater200 & \textbf{DexGrasp Anything + Filter} \\ 
\bottomrule
\end{tabular}
}
\vspace{-2ex}
\label{tab:datasets}
\end{table*}

The quality, diversity and scale of datasets are crucial for advancing dexterous grasping research, especially for diffusion-based generative methods. Training on a broader data distribution enables models to learn richer and more adaptable grasping strategies for arbitrary object. To inspire potential of methods towards universal dexterous grasping, we have developed a comprehensive dataset that significantly exceeds existing dexterous grasping datasets in both size and diversity. In the following sections, we provide a detailed overview of the data construction process, present key statistics, and highlight the characteristics of our \textbf{DexGrasp Anything (DGA) dataset}.

\subsection{Data Construction}\label{sec:data-construction}
Our data construction process begins with curating existing datasets from diverse sources. We gather three simulated datasets~\cite{gendexgrasp-multidex,xu2023unidexgrasp,wang2023dexgraspnet}, a real-world dataset~\cite{liu2024realdex} collected by human operator, alongside GRAB~\cite{taheri2020grab}, a large-scale human hand dataset, to maximize data diversity and richness. Leveraging advancements in robot teleoperation systems like AnyTeleop~\cite{qin2023anyteleop}, we retarget the human hand dataset GRAB to dexterous hand parameters, creating DexGRAB, and filter it to retain only frames with hand-object contact. 
Next, we examine all collected data within IsaacGym~\cite{makoviychuk2021isaac}, applying strict conditions to ensure stability and contact integrity. Specifically, we enforce that \textit{(1) objects do not shift more than 2 cm in any direction under force} and that \textit{(2) hand-object penetration remains below 10 mm and object-hand penetration remains below 1 mm following}~\cite{wang2023dexgraspnet}. The detailed process for computing the penetration is provided in the supplementary materials. This rigorous filtering process guarantees consistent high quality across all data sources.

Training our physics-aware diffusion generator on this dataset leads to higher success rates, greater diversity, and faster generation speeds for zero-shot dexterous grasping on unseen objects. Acting as a data engine, our model facilitates further dataset expansion in a ``model-in-the-loop'' manner. We meticulously selected object meshes from the Objaverse~\cite{deitke2023objaverse, deitke2024objaverse} dataset, with the goal of ensuring broad category coverage and maintaining an even distribution across these categories. To achieve this, we examined all objects within 18 chosen categories and ultimately selected 10,034 distinct objects, covering 6,994 unique tags in the Objaverse data configuration. 
We apply approximate convex decomposition~\cite{wei2022approximate} to each mesh to reduce complexity and ensure water-tightness. Our trained generator then iteratively produces dexterous grasp poses, which are filtered under the same stringent standards. Finally, we combine the curated and generated data to form a large-scale and diverse dataset, crafted to advance research in dexterous grasping.

\begin{table*}[ht]
\caption{Performance comparison across different methods and datasets. Bold numbers indicate the best scores, while underlined numbers indicate the second-best scores. DexGrasp Anything (w/ LLM) achieves the highest or near-highest performance across most metrics.}
\centering
\resizebox{\linewidth}{!}{
\begin{tabular}{@{}l@{\hspace{0.2cm}}*{20}{@{\hspace{0.1cm}}c}@{}}
\toprule
\multirow{2}{*}{\hspace{0.2cm} \makecell{\centering \diagbox[width=2.5cm, height=0.9cm, linewidth=0.8pt]{\textbf{Method}}{\textbf{Dataset}}}} & \multicolumn{4}{c}{DexGraspNet} & \multicolumn{4}{c}{UniDexGrasp} & \multicolumn{4}{c}{MultiDex} & \multicolumn{4}{c}{RealDex} & \multicolumn{4}{c}{DexGRAB} \\ 
\cmidrule(lr){2-5} \cmidrule(lr){6-9} \cmidrule(lr){10-13} \cmidrule(lr){14-17} \cmidrule(lr){18-21}
&\textbf{Suc.6 $\uparrow$} & \textbf{Suc.1 $\uparrow$} & \textbf{Pen. $\downarrow$} & \textbf{Div $\uparrow$} &\textbf{Suc.6 $\uparrow$} & \textbf{Suc.1 $\uparrow$} & \textbf{Pen. $\downarrow$} & \textbf{Div $\uparrow$}& \textbf{Suc.6 $\uparrow$} & \textbf{Suc.1 $\uparrow$} & \textbf{Pen. $\downarrow$} & \textbf{Div $\uparrow$}& \textbf{Suc.6 $\uparrow$} & \textbf{Suc.1 $\uparrow$} & \textbf{Pen. $\downarrow$} & \textbf{Div $\uparrow$}& \textbf{Suc.6 $\uparrow$} & \textbf{Suc.1 $\uparrow$} & \textbf{Pen. $\downarrow$} & \textbf{Div $\uparrow$} \\ \midrule
UniDexGrasp~\cite{xu2023unidexgrasp}  & 33.9 & 70.1 & 31.9 & 0.14 & 23.7 & 65.5 & 24.5 & 0.14 & 21.6 & 47.5 & 13.5 & 0.08 & 27.1 & 59.4 & 39.0 & 0.11 & 20.8 & 55.8 & 37.4 & 0.08 \\
GraspTTA~\cite{grasptta}  & 18.6 & 67.8 & 24.5 & 0.13 & 21.0 & 65.3 & 21.2 & 0.10 & 30.3 & 62.8 & 19.0 & 0.11 & 13.3 & 46.4 & 40.1 & 0.09 & 14.4 & 51.0 & 51.4 & 0.10 \\
SceneDiffuser~\cite{scenediffuser}  & 26.6 & 66.9 & 31.0 & 0.15 & 28.3 & 74.8 & 25.1 & 0.15 & 69.8 & 85.6 & 14.6 & \textbf{0.27} & 21.7 & 56.1 & 42.0 & 0.09 & 39.1 & 85.0 & 41.1 & \underline{0.12} \\
UGG~\cite{lu2023ugg}  & 46.9 & 79.0 & 25.2 & 0.14 & 46.0 & 83.2 & 24.5 & 0.14 & 55.3 & 93.4 & \underline{10.3} & 0.12 & 32.7& 63.4 & 34.4 & 0.10 & 42.7 & 90.6 & 33.2 & \underline{0.12} \\ \midrule
Ours & \underline{53.6} & \underline{90.4} & \underline{21.5} & \underline{0.22} & \textbf{54.8} & \underline{90.8} & \underline{18.9} & \textbf{0.25} & \underline{72.2} & \underline{96.3} & \textbf{9.6} & \underline{0.23} & \underline{34.6} & \underline{71.2} & \textbf{23.1} & \textbf{0.14} & \underline{56.5} & \underline{91.8} & \textbf{28.6} & \underline{0.12} \\
Ours(w/ LLM)  & \textbf{57.5} & \textbf{90.6} & \textbf{17.8} & \textbf{0.23} & \underline{53.1} & \textbf{91.2} & \textbf{18.8} & \underline{0.23} & \textbf{79.1} & \textbf{98.1} & 11.4 & 0.22 & \textbf{44.8} & \textbf{73.7} & \underline{27.7} & \underline{0.13} & \textbf{57.9} & \textbf{92.7} & \underline{30.4} & \textbf{0.13} \\
\bottomrule
\end{tabular}}
\label{tab:comparison}
\end{table*}


\subsection{Statistics}\label{sec:statistics}
We present a comparative analysis of key metrics between our dataset and other existing datasets in Table~\ref{tab:datasets}. Our DGA dataset comprises two main components:  The first component \textbf{DGA-curated} includes approximately 0.88 million grasping poses across 5,664 distinct objects, curated from various existing and diverse data sources. 
The second component \textbf{DGA-generated} is generated with our DexGrasp Anything generator from the Objaverse~\cite{deitke2023objaverse,deitke2024objaverse} dataset, containing approximately 2.52 million grasping poses spanning 10,034 different objects, covering 6,994 unique tags.
In total, our dataset features over 3.4 million grasping poses across 15,698 objects from diverse data distribution, supporting in-depth research into dexterous grasping.

\subsection{Characteristics}\label{sec:characteristics}
\begin{figure}[t]
  \centering
   \includegraphics[width=0.9\linewidth]{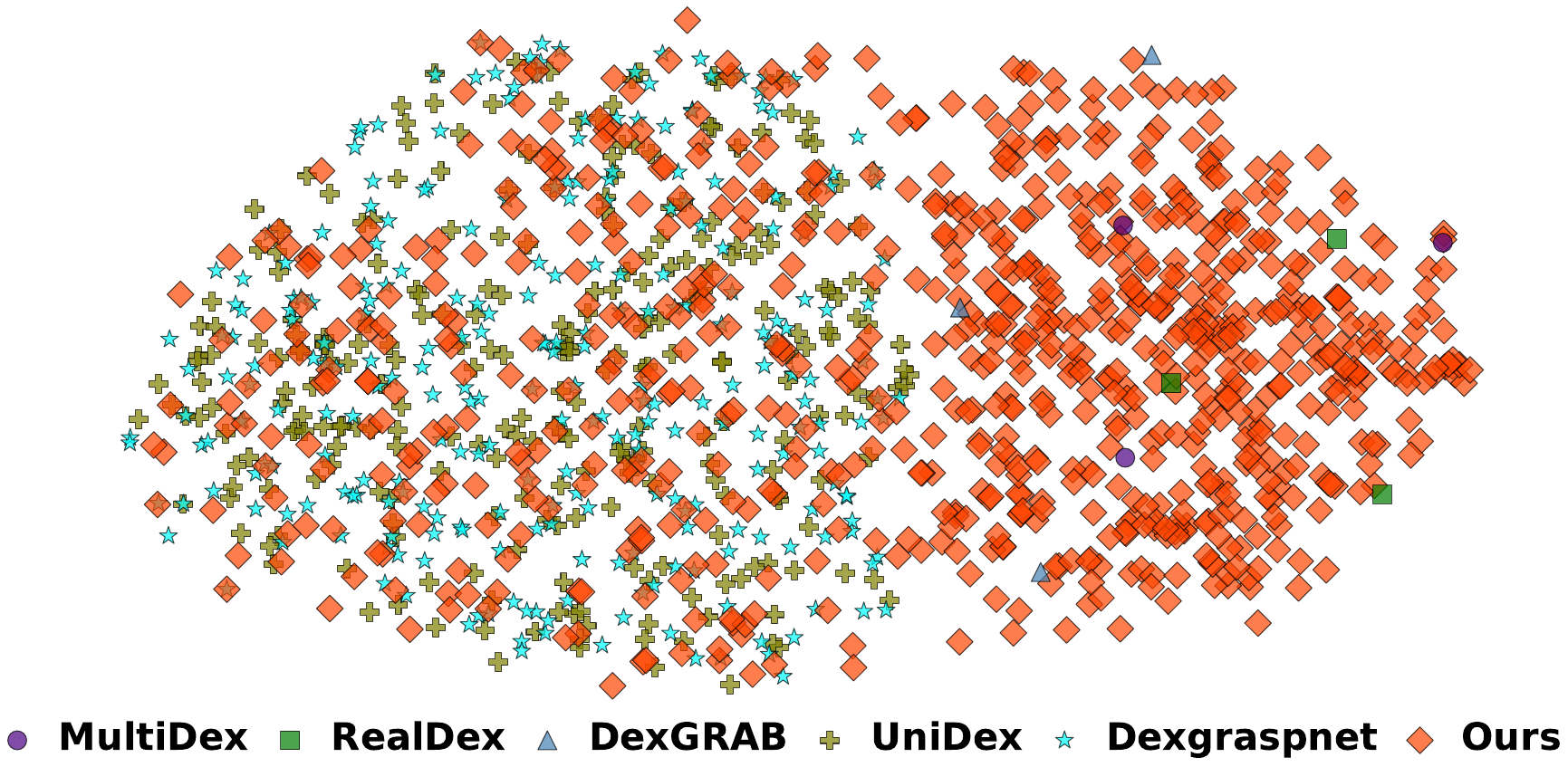}
    \caption{t-SNE visualization of the object features in our dataset compared to existing datasets. Each point represents an object, and different markers and colors are used to distinguish between datasets. For clarity, we randomly sample 5\% objects from each dataset for visualization. 
    }
   \label{fig:object-distribution}
   \vspace{-3ex}
\end{figure}
Our dataset is characterized by a high degree of diversity and comprehensiveness, aimed at capturing a wide range of object and pose variations to advance the performance of dexterous hands in complex real-world environment. The key characteristics of our dataset can be presented as follow:
\begin{itemize}
 \item \textbf{Large data scale.} Our dataset features over 3.4M strictly-tested grasping poses, which is significantly larger than all previous datasets. 
    \item \textbf{Diverse objects.} Our dataset encompasses 15,698 objects from a wide range of categories and sources, ensuring a high level of diversity. In Figure~\ref{fig:object-distribution}, we present a t-SNE visualization comparing object features from our dataset with those from existing datasets, using features extracted by a pre-trained Point Transformer. Object features from our dataset spread much wider across the feature space, suggesting that our dataset captures a broader variety or unique features that not be as present in existing datasets.
    \item \textbf{Diverse grasping poses.} The wide variety of objects contributes to a diverse range of grasping poses. Extensive experimental results in Sec.~\ref{sec:exp-dataset} demonstrate that our dataset significantly enhances the diversity of outcomes from existing methods, while maintaining or even improving the grasping success rate.
\end{itemize}

\section{Experiments}
\label{sec:exp}

\subsection{Comparison}

\textbf{Metrics.} Following previous works~\cite{scenediffuser}, we assess the grasping success rate (Suc.~6/Suc.~1) and maximum penetration (Pen.) in millimeters to gauge the quality of generated poses. A grasping pose is considered successful if the object’s displacement, when an external force is applied, is less than 2 cm in at least one (Suc.~1) or all (Suc.~6) of the six axis-aligned directions in a 3D coordinate system. 
Additionally, we evaluate the diversity using the mean standard deviation of the pose parameters (Div.) across successful grasps in millimeters. All poses are evaluated in the IssacGym~\cite{makoviychuk2021isaac} simulator with the same configuration used in \cite{scenediffuser}. 


\noindent \textbf{Implementation Details.} 
Following \cite{scenediffuser}, we employ a U-Net~\cite{ronneberger2015unet} structure for our diffusion backbone.  An object-conditioned Point Transformer~\cite{zhao2021point-transformer} encoder handles the point clouds, injected into the diffusion model using a cross-attention mechanism.  All point clouds are downsampled to 2048 points before encoding. Our model is implemented using the PyTorch~\cite{NEURIPS2019_9015} platform, optimized with the Adam~\cite{kingma2014adam} algorithm at a learning rate of 0.0001. 
We follow the official train-test split of all datasets. All the compared methods are trained and inferred following their official code implementations. Training and evaluation are carried out on a Linux server equipped with four NVIDIA Tesla A40 GPUs until convergence.

\begin{figure}[]
  \centering
   \includegraphics[width=1.0\linewidth]{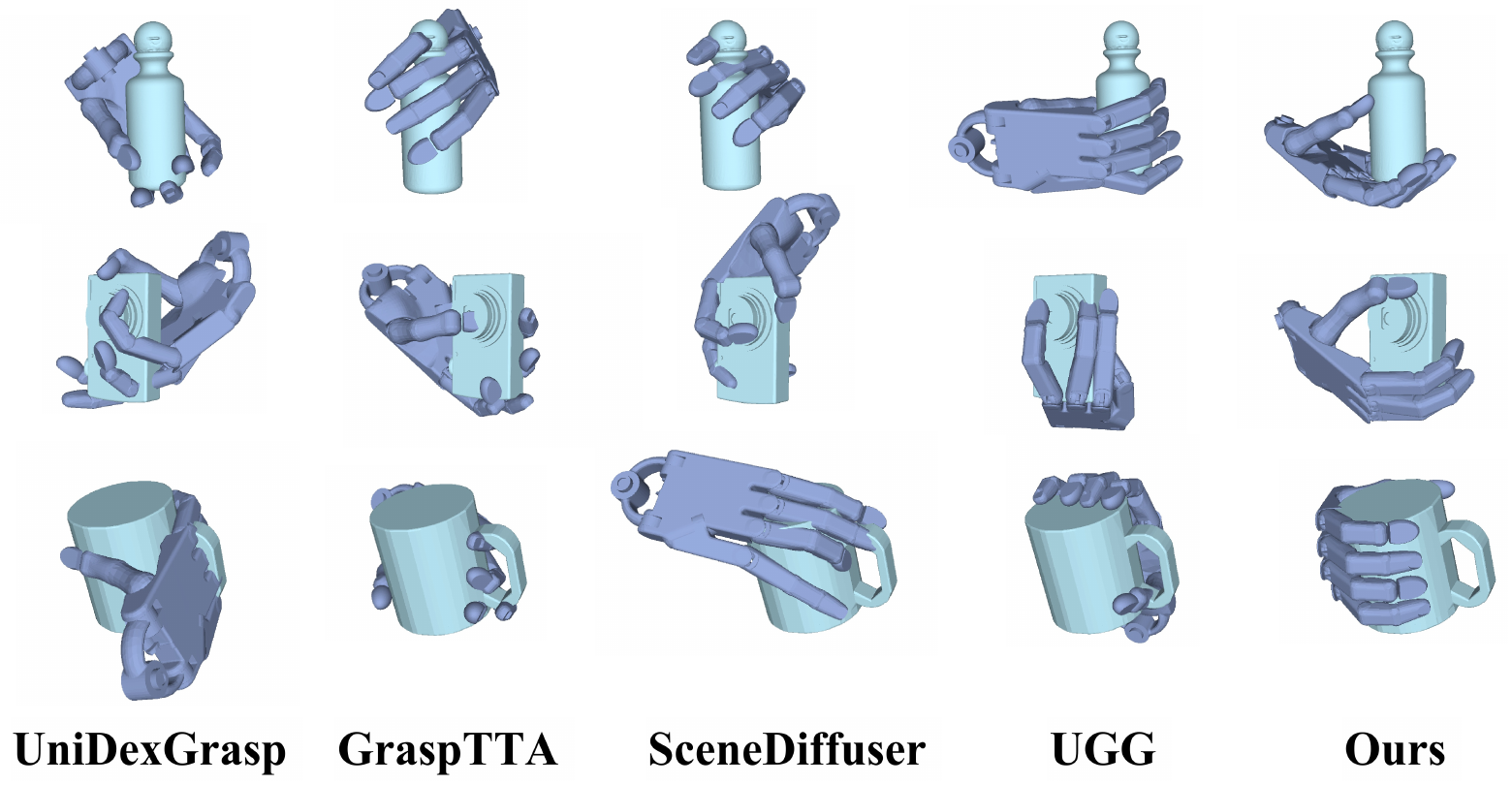}
    \caption{Qualitative visualization of grasping results in Table 2.}
   \label{fig:table2_visual}
\end{figure}

\noindent \textbf{Results.} Table~\ref{tab:comparison} presents the quantitative comparisons. Our method, leveraging a physics-aware training paradigm and a physics-guided sampler, demonstrates superior performance in both pose quality (Suc.~1, Suc.~6, and Pen.) and diversity (Div.) compared to previous methods across all five benchmarks. Qualitative results, shown in Figure ~\ref{fig:table2_visual}, further illustrate that our approach produces more accurate grasping poses, benefiting from the effective physical constraints introduced in both training and sampling stages.

\begin{table}[]
\caption{Ablation study on incorporating physical constraints during both training and sampling stages and the LLM module. The evaluation is conducted on  the DexGraspNet dataset.}
\centering
\resizebox{\linewidth}{!}{
\begin{tabular}{c|cccc|cccc} 
\toprule
 & \textbf{SRF} & \textbf{ERF} & \textbf{SPF} & \textbf{LLM} & \textbf{Suc.6 $\uparrow$} & \textbf{Suc.1 $\uparrow$} & \textbf{Pen. $\downarrow$} & \textbf{Div $\uparrow$} \\ 
\midrule
a &   &   &   &   & 26.6 & 66.9 & 31.0 & 0.15 \\
b & \checkmark &   &   &   & 38.9 & 78.4 & 27.0 & 0.03 \\
c & \checkmark & \checkmark & &  & 46.6 & 83.4  & 15.8 & 0.22   \\
d & \checkmark & \checkmark & \checkmark &   & 53.6 & 90.4 & 21.5 & 0.22 \\
e & \checkmark & \checkmark & \checkmark & \checkmark & 57.5 & 90.6 & 17.8 & 0.23 \\
\midrule
f & \multicolumn{4}{c|}{Constraints only in training}  & 46.8  & 84.0 & 20.9 & 0.18 \\
\bottomrule
\end{tabular}}
\label{tab:ablation_study}
\end{table}

\begin{figure}[]
  \centering
   \includegraphics[width=1.0\linewidth]{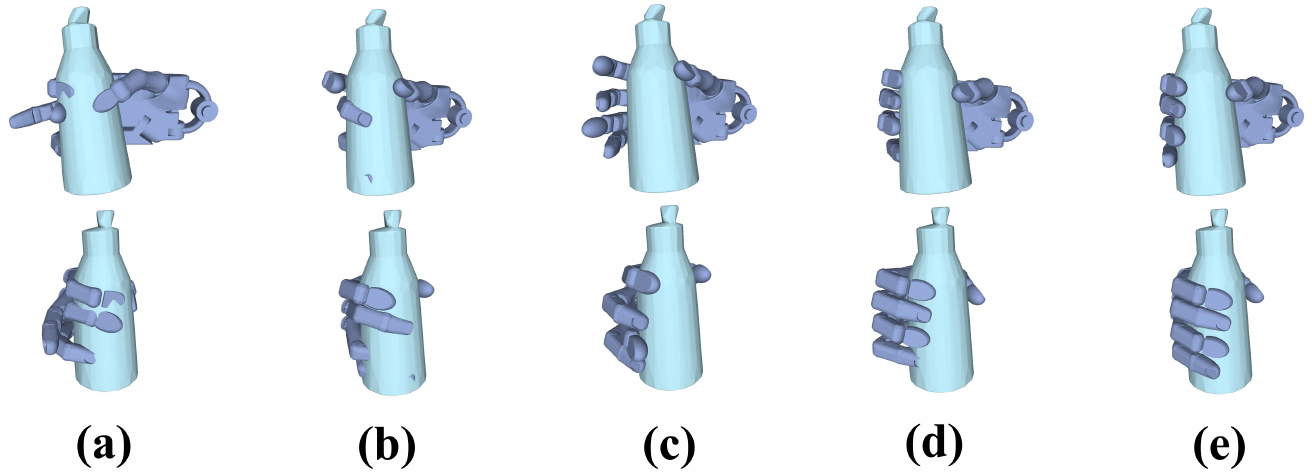}
   \caption{Visualization of the ablation study.  Two rows show different views of each grasp.}
   \vspace{-3ex}
   \label{fig:ablation}
\end{figure}

\subsection{Ablation Study}
In this section, we conduct ablation studies to evaluate the contributions of the proposed physical constraint objectives as well as the LLM-enhanced representation in our DexGrasp Anything generator. These studies are performed on the testing set of the DexGraspNet dataset. The quantitative results are presented in Table~\ref{tab:ablation_study}, where where we incrementally add the three physical constraints and the LLM enhancement (lines a-e) and compare with a model that uses only physical-aware training, excluding the physical-guided sampler (line f). Qualitative results are shown in Figure~\ref{fig:ablation}, with additional samples provided in the supplementary materials. These analyses emphasize the crucial role of each physical constraint, the LLM enhancement, as well as the physics-aware training paradigm and physics-guided sampler in enhancing the system's overall performance.

\subsection{Evaluation for DexGrasp Anything Dataset}\label{sec:exp-dataset}
By gathering high-quality data from various existing sources and augmenting it with our physics-aware diffusion generator, we have constructed the largest and most diverse dexterous grasping dataset to date. This not only enhances the performance of our generator but also benefits other dexterous grasp generation methods.

\noindent \textbf{Setup.} We train our DexGrasp Anything generator, SceneDiffuser, UGG, and GraspTTA on the training sets of both the DexGraspNet and DexGrasp Anything datasets. We evaluate the models on the testing sets of DexGraspNet and RealDex datasets.

\begin{table}[]
\caption{Evaluating Dataset Quality and Cross-Dataset Generalization. Model performance is compared on DexGraspNet and RealDex, with training on either DexGraspNet or our dataset. The best result within each group is highlighted in \textbf{bold}.}
\centering
\resizebox{\linewidth}{!}{
\begin{tabular}{@{}l|@ {\hspace{0.1cm}}l| @{\hspace{0.1cm}}*{8}{@{\hspace{0.1cm}}c}@{}}  
\toprule
\multirow{2}{*}{\textbf{Method}} & \multirow{2}{*}{\hspace{0.2cm} \makecell{\centering \diagbox[width=1.8cm, height=0.9cm, linewidth=0.8pt]{\textbf{Train}}{\textbf{Test}}}} & \multicolumn{4}{c}{DexGraspNet} & \multicolumn{4}{c}{RealDex} \\ 
\cmidrule(lr){3-6} \cmidrule(lr){7-10}
&  & \textbf{Suc.6 $\uparrow$} & \textbf{Suc.1 $\uparrow$} & \textbf{Pen. $\downarrow$} & \textbf{Div $\uparrow$} & \textbf{Suc.6 $\uparrow$} & \textbf{Suc.1 $\uparrow$} & \textbf{Pen. $\downarrow$} & \textbf{Div $\uparrow$} \\ 
\midrule
\multirow{2}{*}{SceneDiffuser} & DexGraspNet &  26.6 & 66.9 & 31.0 & 0.15 & 16.1 & 52.1 & \textbf{29.2} & 0.13 \\
 & DGA dataset & \textbf{40.7} & \textbf{70.6} & \textbf{22.2} & \textbf{0.36} & \textbf{24.5} & \textbf{57.1} & 31.0 & \textbf{0.27}  \\
\midrule
\multirow{2}{*}{GraspTTA} & DexGraspNet & 18.6 & 67.8 & 24.5 & 0.13 & 25.5 & 64.8 & \textbf{31.6} & 0.11 \\
 & DGA dataset & \textbf{28.0} & \textbf{73.0} & \textbf{23.9} & \textbf{0.21} & \textbf{32.9} & \textbf{76.2} & 32.1 & \textbf{0.20}  \\
\midrule
\multirow{2}{*}{UGG} & DexGraspNet & 46.9 & 79.0 & \textbf{25.2} & 0.14 & 33.6 & 74.5 & \textbf{33.0} & 0.13 \\
 & DGA dataset & \textbf{49.1} & \textbf{85.9} & 26.2 & \textbf{0.22} & \textbf{42.9} & \textbf{77.3} & 34.4 & \textbf{0.22} \\
\midrule
\multirow{3}{*}{Ours} &DexGraspNet & 53.6 & \textbf{90.4} & 21.5 & 0.22 & 38.4 & 77.5 & \textbf{19.2} & 0.17  \\
 & DGA-curated & 55.9 & 87.3 & 20.9 & 0.28& 52.6 & \textbf{85.7} & 21.5 & 0.25\\
 & DGA dataset & \textbf{58.6} & 88.5 & \textbf{17.8} & \textbf{0.38} & \textbf{53.4} & 84.4 & 22.4 & \textbf{0.32}  \\
\bottomrule
\end{tabular}}
\label{tab:joint-training}
\end{table}

\begin{figure}[ht]
  \centering
   \includegraphics[width=1.0\linewidth]{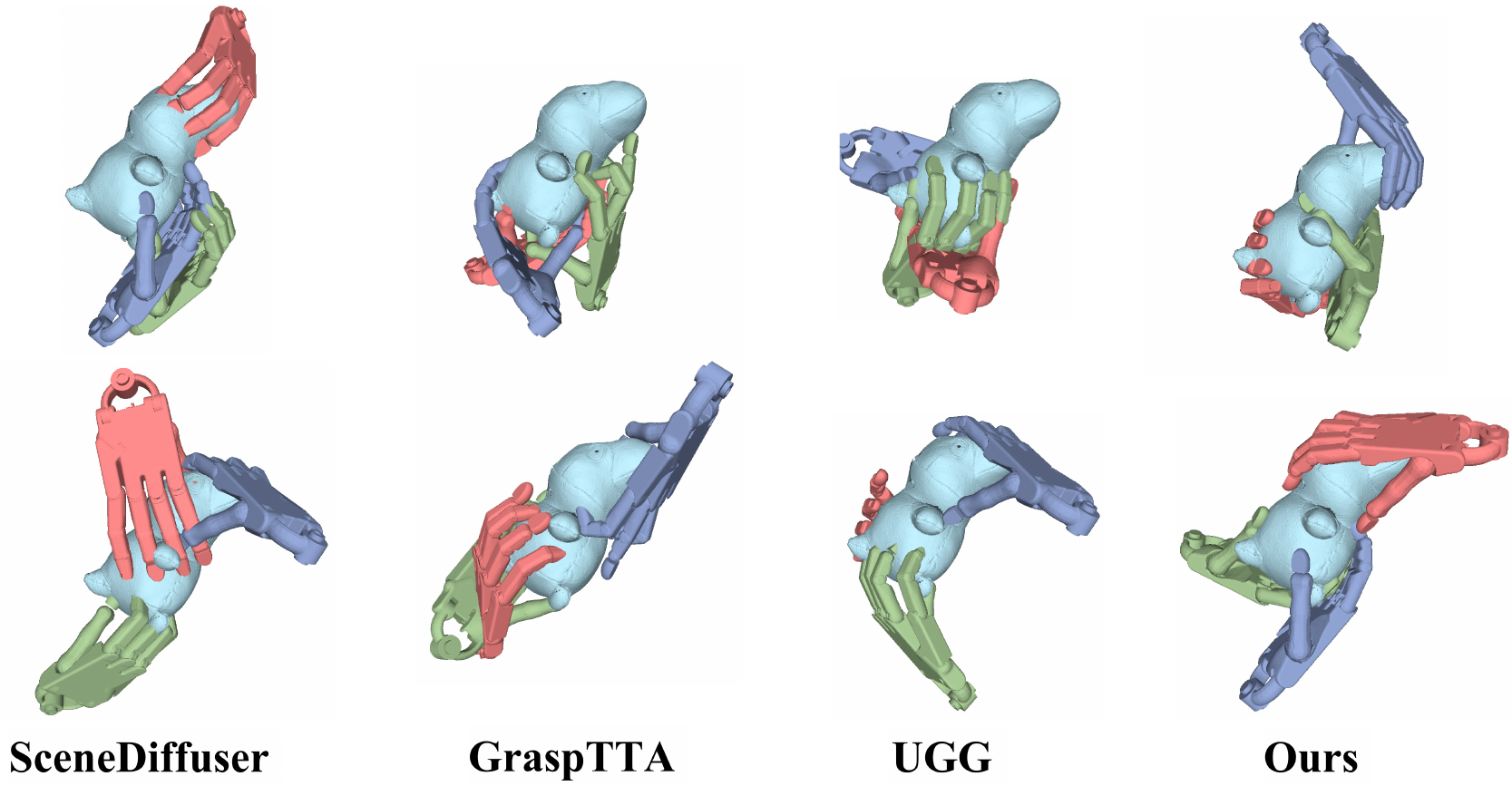}

   \caption{Visualization of cross-dataset evaluation results shown in Table 4. The top row shows models trained on DexGraspNet, while the bottom row displays models trained on our dataset.}
   \label{fig:joint-training}
\end{figure}
\noindent \textbf{Results.} As shown in Table~\ref{tab:joint-training}, our findings indicate that training on the DexGrasp Anything dataset significantly improves the diversity of the sampling results for DexGrasp Anything generator, UGG, SceneDiffuser and GraspTTA. This is achieved without compromising, and in some cases improving, the grasping success rate and penetration metrics. To further validate our approach, we provide a more comprehensive evaluation in the supplementary materials. 
The qualitative comparison results are illustrated in Figure~\ref{fig:joint-training}. For a variety of target objects, models trained on the DGA dataset generate significantly more diverse grasping poses, while maintaining high-quality results. 
\begin{figure}[t]
  \centering
   \includegraphics[width=1.0\linewidth]{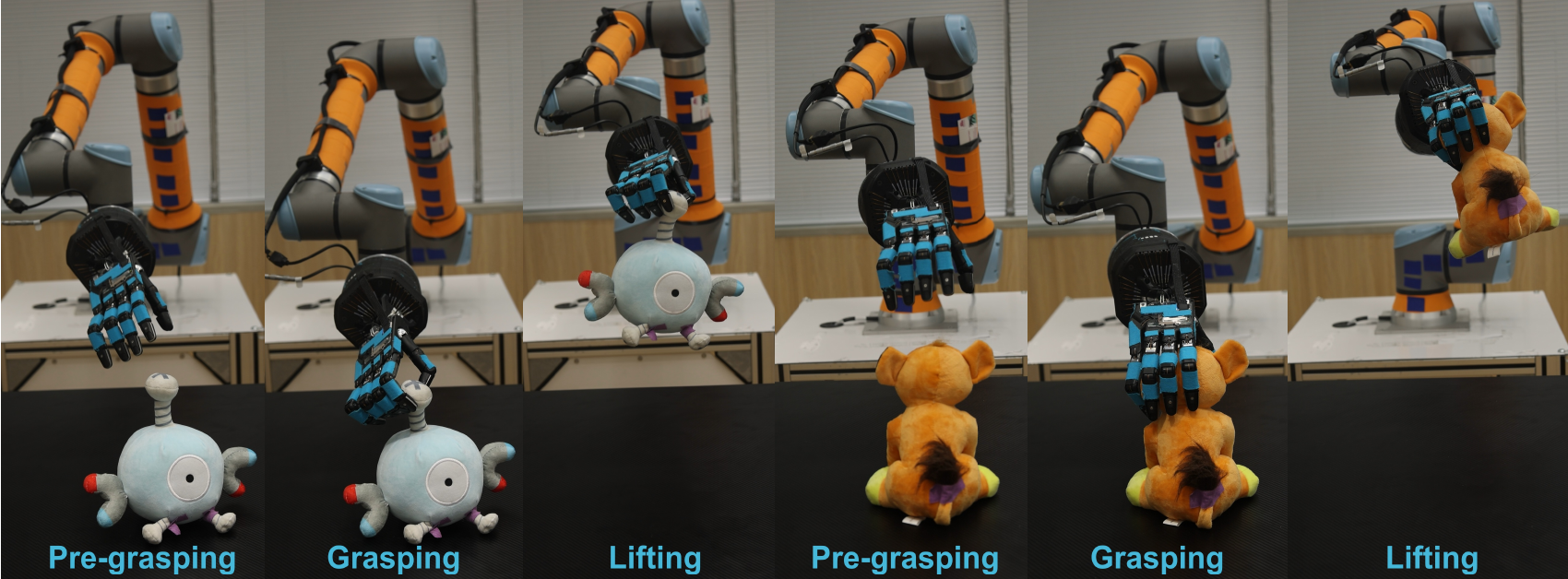}

   \caption{Real-world evaluation for our method.}
   \vspace{-6mm}
   \label{fig:grasping motion sequence}
\end{figure}

\subsection{Real-world Application}
Harnessing physics and semantic priors, and trained on our diverse, high-quality dataset, DexGrasp Anything is highly capable of producing robust and practicable grasping poses in real-world environments. To validate its performance, we deploy the model on a real ShadowHand robot, as shown in Figure~\ref{fig:grasping motion sequence}. The pre-grasping motion sequence is generated following the approach in~\cite{liu2024realdex}. The real-world experiments demonstrate that our grasping motions are reasonable and stable for unseen real objects, proving the universality and practicability of our method. More video demos are available in the supplementary materials.

\section{Conclusions}
\label{sec:conclusions}
We introduce DexGrasp Anything, a physics-aware diffusion generator designed for universal and robust dexterous grasp generation. It deeply incorporates three tailored physical constraints through a physics-aware training paradigm and a physics-guided sampler. Moreover, we present the largest and most diverse dataset for dexterous grasp generation to date. Extensive experiments demonstrate that our method and dataset significantly enhance the quality and diversity of dexterous grasp generation. We believe our contributions will pave the way for future advancements towards the universal robotic dexterous grasping.


{
\small
\bibliographystyle{ieeenat_fullname}
\bibliography{main}

\begin{thebibliography}{52}
\providecommand{\natexlab}[1]{#1}
\providecommand{\url}[1]{\texttt{#1}}
\expandafter\ifx\csname urlstyle\endcsname\relax
  \providecommand{\doi}[1]{doi: #1}\else
  \providecommand{\doi}{doi: \begingroup \urlstyle{rm}\Url}\fi

\bibitem[Chung et~al.(2022)Chung, Kim, McCann, Klasky, and Ye]{Chung2022DiffusionPS}
Hyungjin Chung, Jeongsol Kim, Michael~T. McCann, Marc~Louis Klasky, and J.~C. Ye.
\newblock Diffusion posterior sampling for general noisy inverse problems.
\newblock \emph{ArXiv}, abs/2209.14687, 2022.

\bibitem[Deitke et~al.(2023)Deitke, Schwenk, Salvador, Weihs, Michel, VanderBilt, Schmidt, Ehsani, Kembhavi, and Farhadi]{deitke2023objaverse}
Matt Deitke, Dustin Schwenk, Jordi Salvador, Luca Weihs, Oscar Michel, Eli VanderBilt, Ludwig Schmidt, Kiana Ehsani, Aniruddha Kembhavi, and Ali Farhadi.
\newblock Objaverse: A universe of annotated 3d objects.
\newblock In \emph{Proceedings of the IEEE/CVF Conference on Computer Vision and Pattern Recognition}, pages 13142--13153, 2023.

\bibitem[Deitke et~al.(2024)Deitke, Liu, Wallingford, Ngo, Michel, Kusupati, Fan, Laforte, Voleti, Gadre, et~al.]{deitke2024objaverse}
Matt Deitke, Ruoshi Liu, Matthew Wallingford, Huong Ngo, Oscar Michel, Aditya Kusupati, Alan Fan, Christian Laforte, Vikram Voleti, Samir~Yitzhak Gadre, et~al.
\newblock Objaverse-xl: A universe of 10m+ 3d objects.
\newblock \emph{Advances in Neural Information Processing Systems}, 36, 2024.

\bibitem[Dhariwal and Nichol(2021)]{Dhariwal2021Diffusionbeatgan}
Prafulla Dhariwal and Alex Nichol.
\newblock Diffusion models beat gans on image synthesis.
\newblock \emph{ArXiv}, abs/2105.05233, 2021.

\bibitem[Ferrari et~al.(1992)Ferrari, Canny, et~al.]{ap1}
Carlo Ferrari, John~F Canny, et~al.
\newblock Planning optimal grasps.
\newblock In \emph{ICRA}, page~6, 1992.

\bibitem[Hasson et~al.(2019)Hasson, Varol, Tzionas, Kalevatykh, Black, Laptev, and Schmid]{graspit-data2}
Yana Hasson, Gul Varol, Dimitrios Tzionas, Igor Kalevatykh, Michael~J Black, Ivan Laptev, and Cordelia Schmid.
\newblock Learning joint reconstruction of hands and manipulated objects.
\newblock In \emph{Proceedings of the IEEE/CVF conference on computer vision and pattern recognition}, pages 11807--11816, 2019.

\bibitem[Ho(2022)]{Ho2022ClassifierFreeDG}
Jonathan Ho.
\newblock Classifier-free diffusion guidance.
\newblock \emph{ArXiv}, abs/2207.12598, 2022.

\bibitem[Ho et~al.(2020)Ho, Jain, and Abbeel]{ddpm}
Jonathan Ho, Ajay Jain, and Pieter Abbeel.
\newblock Denoising diffusion probabilistic models.
\newblock \emph{Advances in neural information processing systems}, 33:\penalty0 6840--6851, 2020.

\bibitem[Huang et~al.(2023{\natexlab{a}})Huang, Huang, Yang, Ren, Liu, Li, Ye, Liu, Yin, and Zhao]{Huang2023MakeAnAudioTG}
Rongjie Huang, Jia-Bin Huang, Dongchao Yang, Yi Ren, Luping Liu, Mingze Li, Zhenhui Ye, Jinglin Liu, Xiaoyue Yin, and Zhou Zhao.
\newblock Make-an-audio: Text-to-audio generation with prompt-enhanced diffusion models.
\newblock \emph{ArXiv}, abs/2301.12661, 2023{\natexlab{a}}.

\bibitem[Huang et~al.(2023{\natexlab{b}})Huang, Wang, Li, Jia, Liu, Zhu, Liang, and Zhu]{scenediffuser}
Siyuan Huang, Zan Wang, Puhao Li, Baoxiong Jia, Tengyu Liu, Yixin Zhu, Wei Liang, and Song-Chun Zhu.
\newblock Diffusion-based generation, optimization, and planning in 3d scenes.
\newblock In \emph{Proceedings of the IEEE/CVF Conference on Computer Vision and Pattern Recognition}, pages 16750--16761, 2023{\natexlab{b}}.

\bibitem[Jiang et~al.(2021)Jiang, Liu, Wang, and Wang]{grasptta}
Hanwen Jiang, Shaowei Liu, Jiashun Wang, and Xiaolong Wang.
\newblock Hand-object contact consistency reasoning for human grasps generation.
\newblock In \emph{Proceedings of the IEEE/CVF international conference on computer vision}, pages 11107--11116, 2021.

\bibitem[Karras et~al.(2022)Karras, Aittala, Aila, and Laine]{karras2022elucidating}
Tero Karras, Miika Aittala, Timo Aila, and Samuli Laine.
\newblock Elucidating the design space of diffusion-based generative models.
\newblock \emph{Advances in neural information processing systems}, 35:\penalty0 26565--26577, 2022.

\bibitem[Kingma(2014)]{kingma2014adam}
Diederik~P Kingma.
\newblock Adam: A method for stochastic optimization.
\newblock \emph{arXiv preprint arXiv:1412.6980}, 2014.

\bibitem[Kirillov et~al.(2023)Kirillov, Mintun, Ravi, Mao, Rolland, Gustafson, Xiao, Whitehead, Berg, Lo, et~al.]{kirillov2023sam}
Alexander Kirillov, Eric Mintun, Nikhila Ravi, Hanzi Mao, Chloe Rolland, Laura Gustafson, Tete Xiao, Spencer Whitehead, Alexander~C Berg, Wan-Yen Lo, et~al.
\newblock Segment anything.
\newblock In \emph{Proceedings of the IEEE/CVF International Conference on Computer Vision}, pages 4015--4026, 2023.

\bibitem[Li et~al.(2023)Li, Liu, Li, Geng, Zhu, Yang, and Huang]{gendexgrasp-multidex}
Puhao Li, Tengyu Liu, Yuyang Li, Yiran Geng, Yixin Zhu, Yaodong Yang, and Siyuan Huang.
\newblock Gendexgrasp: Generalizable dexterous grasping.
\newblock In \emph{2023 IEEE International Conference on Robotics and Automation (ICRA)}, pages 8068--8074. IEEE, 2023.

\bibitem[Liu et~al.(2023)Liu, Chen, Yuan, Mei, Liu, Mandic, Wang, and Plumbley]{Liu2023AudioLDMTG}
Haohe Liu, Zehua Chen, Yiitan Yuan, Xinhao Mei, Xubo Liu, Danilo~P. Mandic, Wenwu Wang, and MarkD~. Plumbley.
\newblock Audioldm: Text-to-audio generation with latent diffusion models.
\newblock In \emph{International Conference on Machine Learning}, 2023.

\bibitem[Liu et~al.(2020)Liu, Pan, Xu, Ganguly, and Manocha]{ddg}
Min Liu, Zherong Pan, Kai Xu, Kanishka Ganguly, and Dinesh Manocha.
\newblock Deep differentiable grasp planner for high-dof grippers.
\newblock \emph{arXiv preprint arXiv:2002.01530}, 2020.

\bibitem[Liu et~al.(2021)Liu, Liu, Jiao, Zhu, and Zhu]{ap2}
Tengyu Liu, Zeyu Liu, Ziyuan Jiao, Yixin Zhu, and Song-Chun Zhu.
\newblock Synthesizing diverse and physically stable grasps with arbitrary hand structures using differentiable force closure estimator.
\newblock \emph{IEEE Robotics and Automation Letters}, 7\penalty0 (1):\penalty0 470--477, 2021.

\bibitem[Liu et~al.(2024{\natexlab{a}})Liu, Yang, Wang, Wu, Wang, Yao, Schwertfeger, Yang, Wang, Yu, et~al.]{liu2024realdex}
Yumeng Liu, Yaxun Yang, Youzhuo Wang, Xiaofei Wu, Jiamin Wang, Yichen Yao, S{\"o}ren Schwertfeger, Sibei Yang, Wenping Wang, Jingyi Yu, et~al.
\newblock Realdex: Towards human-like grasping for robotic dexterous hand.
\newblock \emph{arXiv preprint arXiv:2402.13853}, 2024{\natexlab{a}}.

\bibitem[Liu et~al.(2024{\natexlab{b}})Liu, Zhang, Li, Yan, Gao, Chen, Yuan, Huang, Sun, Gao, He, and Sun]{Liu2024SoraAR}
Yixin Liu, Kai Zhang, Yuan Li, Zhiling Yan, Chujie Gao, Ruoxi Chen, Zhengqing Yuan, Yue Huang, Hanchi Sun, Jianfeng Gao, Lifang He, and Lichao Sun.
\newblock Sora: A review on background, technology, limitations, and opportunities of large vision models.
\newblock \emph{ArXiv}, abs/2402.17177, 2024{\natexlab{b}}.

\bibitem[Lu et~al.(2023)Lu, Kang, Li, Liu, Yang, Huang, and Hua]{lu2023ugg}
Jiaxin Lu, Hao Kang, Haoxiang Li, Bo Liu, Yiding Yang, Qixing Huang, and Gang Hua.
\newblock Ugg: Unified generative grasping.
\newblock \emph{arXiv preprint arXiv:2311.16917}, 2023.

\bibitem[Lundell et~al.(2021)Lundell, Verdoja, and Kyrki]{lundell2021ddgc}
Jens Lundell, Francesco Verdoja, and Ville Kyrki.
\newblock Ddgc: Generative deep dexterous grasping in clutter.
\newblock \emph{IEEE Robotics and Automation Letters}, 6\penalty0 (4):\penalty0 6899--6906, 2021.

\bibitem[Makoviychuk et~al.(2021)Makoviychuk, Wawrzyniak, Guo, Lu, Storey, Macklin, Hoeller, Rudin, Allshire, Handa, et~al.]{makoviychuk2021isaac}
Viktor Makoviychuk, Lukasz Wawrzyniak, Yunrong Guo, Michelle Lu, Kier Storey, Miles Macklin, David Hoeller, Nikita Rudin, Arthur Allshire, Ankur Handa, et~al.
\newblock Isaac gym: High performance gpu-based physics simulation for robot learning.
\newblock \emph{arXiv preprint arXiv:2108.10470}, 2021.

\bibitem[Miller and Allen(2004)]{miller2004graspit}
Andrew~T Miller and Peter~K Allen.
\newblock Graspit! a versatile simulator for robotic grasping.
\newblock \emph{IEEE Robotics \& Automation Magazine}, 11\penalty0 (4):\penalty0 110--122, 2004.

\bibitem[Murray et~al.(2017)Murray, Li, and Sastry]{ap6}
Richard~M Murray, Zexiang Li, and S~Shankar Sastry.
\newblock \emph{A mathematical introduction to robotic manipulation}.
\newblock CRC press, 2017.

\bibitem[Paszke et~al.(2019)Paszke, Gross, Massa, Lerer, Bradbury, Chanan, Killeen, Lin, Gimelshein, Antiga, Desmaison, Kopf, Yang, DeVito, Raison, Tejani, Chilamkurthy, Steiner, Fang, Bai, and Chintala]{NEURIPS2019_9015}
Adam Paszke, Sam Gross, Francisco Massa, Adam Lerer, James Bradbury, Gregory Chanan, Trevor Killeen, Zeming Lin, Natalia Gimelshein, Luca Antiga, Alban Desmaison, Andreas Kopf, Edward Yang, Zachary DeVito, Martin Raison, Alykhan Tejani, Sasank Chilamkurthy, Benoit Steiner, Lu Fang, Junjie Bai, and Soumith Chintala.
\newblock Pytorch: An imperative style, high-performance deep learning library.
\newblock In \emph{Advances in Neural Information Processing Systems 32}, pages 8024--8035. Curran Associates, Inc., 2019.

\bibitem[Peebles and Xie(2023)]{peebles2023dit}
William Peebles and Saining Xie.
\newblock Scalable diffusion models with transformers.
\newblock In \emph{Proceedings of the IEEE/CVF International Conference on Computer Vision}, pages 4195--4205, 2023.

\bibitem[Ponce et~al.(1993)Ponce, Sullivan, Boissonnat, and Merlet]{ap3}
Jean Ponce, Steve Sullivan, J-D Boissonnat, and J-P Merlet.
\newblock On characterizing and computing three-and four-finger force-closure grasps of polyhedral objects.
\newblock In \emph{[1993] Proceedings IEEE International Conference on Robotics and Automation}, pages 821--827. IEEE, 1993.

\bibitem[Prattichizzo et~al.(2012)Prattichizzo, Malvezzi, Gabiccini, and Bicchi]{ap4}
Domenico Prattichizzo, Monica Malvezzi, Marco Gabiccini, and Antonio Bicchi.
\newblock On the manipulability ellipsoids of underactuated robotic hands with compliance.
\newblock \emph{Robotics and Autonomous Systems}, 60\penalty0 (3):\penalty0 337--346, 2012.

\bibitem[Qin et~al.(2023)Qin, Yang, Huang, Van~Wyk, Su, Wang, Chao, and Fox]{qin2023anyteleop}
Yuzhe Qin, Wei Yang, Binghao Huang, Karl Van~Wyk, Hao Su, Xiaolong Wang, Yu-Wei Chao, and Dieter Fox.
\newblock Anyteleop: A general vision-based dexterous robot arm-hand teleoperation system.
\newblock In \emph{Robotics: Science and Systems}, 2023.

\bibitem[Ramesh et~al.(2022)Ramesh, Dhariwal, Nichol, Chu, and Chen]{Ramesh2022HierarchicalTI}
Aditya Ramesh, Prafulla Dhariwal, Alex Nichol, Casey Chu, and Mark Chen.
\newblock Hierarchical text-conditional image generation with clip latents.
\newblock \emph{ArXiv}, abs/2204.06125, 2022.

\bibitem[Ren et~al.(2023)Ren, Huang, Zeng, Museth, Fidler, and Williams]{Ren2023XCubeL}
Xuanchi Ren, Jiahui Huang, Xiaohui Zeng, Ken Museth, Sanja Fidler, and Francis Williams.
\newblock Xcube (x3): Large-scale 3d generative modeling using sparse voxel hierarchies.
\newblock \emph{ArXiv}, abs/2312.03806, 2023.

\bibitem[Rombach et~al.(2022)Rombach, Blattmann, Lorenz, Esser, and Ommer]{latent-diffusion}
Robin Rombach, Andreas Blattmann, Dominik Lorenz, Patrick Esser, and Bj{\"o}rn Ommer.
\newblock High-resolution image synthesis with latent diffusion models.
\newblock In \emph{Proceedings of the IEEE/CVF conference on computer vision and pattern recognition}, pages 10684--10695, 2022.

\bibitem[Ronneberger et~al.(2015)Ronneberger, Fischer, and Brox]{ronneberger2015unet}
Olaf Ronneberger, Philipp Fischer, and Thomas Brox.
\newblock U-net: Convolutional networks for biomedical image segmentation.
\newblock In \emph{Medical image computing and computer-assisted intervention--MICCAI 2015: 18th international conference, Munich, Germany, October 5-9, 2015, proceedings, part III 18}, pages 234--241. Springer, 2015.

\bibitem[Rosales et~al.(2012)Rosales, Su{\'a}rez, Gabiccini, and Bicchi]{ap5}
Carlos Rosales, Ra{\'u}l Su{\'a}rez, Marco Gabiccini, and Antonio Bicchi.
\newblock On the synthesis of feasible and prehensile robotic grasps.
\newblock In \emph{2012 IEEE international conference on robotics and automation}, pages 550--556. IEEE, 2012.

\bibitem[Saharia et~al.(2022)Saharia, Chan, Saxena, Li, Whang, Denton, Ghasemipour, Gontijo~Lopes, Karagol~Ayan, Salimans, et~al.]{saharia2022imagen}
Chitwan Saharia, William Chan, Saurabh Saxena, Lala Li, Jay Whang, Emily~L Denton, Kamyar Ghasemipour, Raphael Gontijo~Lopes, Burcu Karagol~Ayan, Tim Salimans, et~al.
\newblock Photorealistic text-to-image diffusion models with deep language understanding.
\newblock \emph{Advances in neural information processing systems}, 35:\penalty0 36479--36494, 2022.

\bibitem[Shaw et~al.(2023)Shaw, Bahl, and Pathak]{shaw2023videodex}
Kenneth Shaw, Shikhar Bahl, and Deepak Pathak.
\newblock Videodex: Learning dexterity from internet videos.
\newblock In \emph{Conference on Robot Learning}, pages 654--665. PMLR, 2023.

\bibitem[Sivakumar et~al.(2022)Sivakumar, Shaw, and Pathak]{sivakumar2022robotic}
Aravind Sivakumar, Kenneth Shaw, and Deepak Pathak.
\newblock Robotic telekinesis: Learning a robotic hand imitator by watching humans on youtube.
\newblock \emph{arXiv preprint arXiv:2202.10448}, 2022.

\bibitem[Song et~al.(2020)Song, Sohl-Dickstein, Kingma, Kumar, Ermon, and Poole]{song2020sde}
Yang Song, Jascha Sohl-Dickstein, Diederik~P Kingma, Abhishek Kumar, Stefano Ermon, and Ben Poole.
\newblock Score-based generative modeling through stochastic differential equations.
\newblock \emph{arXiv preprint arXiv:2011.13456}, 2020.

\bibitem[Taheri et~al.(2020)Taheri, Ghorbani, Black, and Tzionas]{taheri2020grab}
Omid Taheri, Nima Ghorbani, Michael~J Black, and Dimitrios Tzionas.
\newblock Grab: A dataset of whole-body human grasping of objects.
\newblock In \emph{Computer Vision--ECCV 2020: 16th European Conference, Glasgow, UK, August 23--28, 2020, Proceedings, Part IV 16}, pages 581--600. Springer, 2020.

\bibitem[Wang et~al.(2023)Wang, Zhang, Chen, Xu, Li, Liu, and Wang]{wang2023dexgraspnet}
Ruicheng Wang, Jialiang Zhang, Jiayi Chen, Yinzhen Xu, Puhao Li, Tengyu Liu, and He Wang.
\newblock Dexgraspnet: A large-scale robotic dexterous grasp dataset for general objects based on simulation.
\newblock In \emph{2023 IEEE International Conference on Robotics and Automation (ICRA)}, pages 11359--11366. IEEE, 2023.

\bibitem[Wei et~al.(2022)Wei, Liu, Ling, and Su]{wei2022approximate}
Xinyue Wei, Minghua Liu, Zhan Ling, and Hao Su.
\newblock Approximate convex decomposition for 3d meshes with collision-aware concavity and tree search.
\newblock \emph{ACM Transactions on Graphics (TOG)}, 41\penalty0 (4):\penalty0 1--18, 2022.

\bibitem[Wei et~al.(2024)Wei, Jiang, Xing, Tan, Wu, Li, Cutkosky, and Zheng]{GYS}
Yi-Lin Wei, Jian-Jian Jiang, Chengyi Xing, Xiantuo Tan, Xiao-Ming Wu, Hao Li, Mark Cutkosky, and Wei-Shi Zheng.
\newblock Grasp as you say: Language-guided dexterous grasp generation.
\newblock \emph{arXiv preprint arXiv:2405.19291}, 2024.

\bibitem[Weng et~al.(2024)Weng, Lu, Kragic, and Lundell]{weng2024dexdiffuser}
Zehang Weng, Haofei Lu, Danica Kragic, and Jens Lundell.
\newblock Dexdiffuser: Generating dexterous grasps with diffusion models.
\newblock \emph{arXiv preprint arXiv:2402.02989}, 2024.

\bibitem[Xu et~al.(2024)Xu, Wei, Zheng, Wu, and Zheng]{DGTR}
Guo-Hao Xu, Yi-Lin Wei, Dian Zheng, Xiao-Ming Wu, and Wei-Shi Zheng.
\newblock Dexterous grasp transformer.
\newblock In \emph{Proceedings of the IEEE/CVF Conference on Computer Vision and Pattern Recognition}, pages 17933--17942, 2024.

\bibitem[Xu et~al.(2022)Xu, Yu, Song, Shi, Ermon, and Tang]{Xu2022GeoDiffAG}
Minkai Xu, Lantao Yu, Yang Song, Chence Shi, Stefano Ermon, and Jian Tang.
\newblock Geodiff: a geometric diffusion model for molecular conformation generation.
\newblock \emph{ArXiv}, abs/2203.02923, 2022.

\bibitem[Xu et~al.(2023)Xu, Wan, Zhang, Liu, Shan, Shen, Wang, Geng, Weng, Chen, et~al.]{xu2023unidexgrasp}
Yinzhen Xu, Weikang Wan, Jialiang Zhang, Haoran Liu, Zikang Shan, Hao Shen, Ruicheng Wang, Haoran Geng, Yijia Weng, Jiayi Chen, et~al.
\newblock Unidexgrasp: Universal robotic dexterous grasping via learning diverse proposal generation and goal-conditioned policy.
\newblock In \emph{Proceedings of the IEEE/CVF Conference on Computer Vision and Pattern Recognition}, pages 4737--4746, 2023.

\bibitem[Yang et~al.(2024)Yang, Ding, Cai, Yu, Wang, and Shi]{Yang2024diffusion-gaussian-constraint}
Lingxiao Yang, Shutong Ding, Yifan Cai, Jingyi Yu, Jingya Wang, and Ye Shi.
\newblock Guidance with spherical gaussian constraint for conditional diffusion.
\newblock \emph{ArXiv}, abs/2402.03201, 2024.

\bibitem[Zhang et~al.(2024{\natexlab{a}})Zhang, Christen, Fan, Hilliges, and Song]{zhang2024graspxl}
Hui Zhang, Sammy Christen, Zicong Fan, Otmar Hilliges, and Jie Song.
\newblock {GraspXL}: Generating grasping motions for diverse objects at scale.
\newblock In \emph{European Conference on Computer Vision (ECCV)}, 2024{\natexlab{a}}.

\bibitem[Zhang et~al.(2024{\natexlab{b}})Zhang, Wang, Zhang, Qiu, Pang, Jiang, Yang, Xu, and Yu]{zhang2024clay}
Longwen Zhang, Ziyu Wang, Qixuan Zhang, Qiwei Qiu, Anqi Pang, Haoran Jiang, Wei Yang, Lan Xu, and Jingyi Yu.
\newblock Clay: A controllable large-scale generative model for creating high-quality 3d assets.
\newblock \emph{ACM Transactions on Graphics (TOG)}, 43\penalty0 (4):\penalty0 1--20, 2024{\natexlab{b}}.

\bibitem[Zhang et~al.(2024{\natexlab{c}})Zhang, Zhou, Liu, Liu, Yuan, Guo, Zhao, Ang~Jr, and Tay]{zhang2024dexgrasp-diffusion}
Zhengshen Zhang, Lei Zhou, Chenchen Liu, Zhiyang Liu, Chengran Yuan, Sheng Guo, Ruiteng Zhao, Marcelo~H Ang~Jr, and Francis~EH Tay.
\newblock Dexgrasp-diffusion: Diffusion-based unified functional grasp synthesis pipeline for multi-dexterous robotic hands.
\newblock \emph{arXiv preprint arXiv:2407.09899}, 2024{\natexlab{c}}.

\bibitem[Zhao et~al.(2021)Zhao, Jiang, Jia, Torr, and Koltun]{zhao2021point-transformer}
Hengshuang Zhao, Li Jiang, Jiaya Jia, Philip~HS Torr, and Vladlen Koltun.
\newblock Point transformer.
\newblock In \emph{Proceedings of the IEEE/CVF international conference on computer vision}, pages 16259--16268, 2021.

\end{thebibliography}
}

\label{sec:appendix}
\clearpage
\appendix

\noindent \textbf{\Large Appendix}

\section{Overview}
In the main text, we introduce DexGrasp Anything, a physics-aware diffusion generator that incorporates three tailored physical constraints for generating dexterous grasps. Along with this, we present the largest and most diverse dataset for dexterous grasp generation to date. To further demonstrate the improvements brought by our method and dataset, this supplementary material provides more comprehensive experimental results (Sec.~\ref{sec: sup-result}) and details the filtering process (Sec.~\ref{sec: details}) used for dataset construction. Additionally, we have included \textbf{a demo video} in the supplementary files that showcases our \textbf{zero-shot real-world experiments on unseen objects}, which we highly recommend reviewing for a deeper understanding of our method’s practical applications and performance.

\begin{figure}[htbp]
  \centering
   \includegraphics[width=1.0\linewidth]{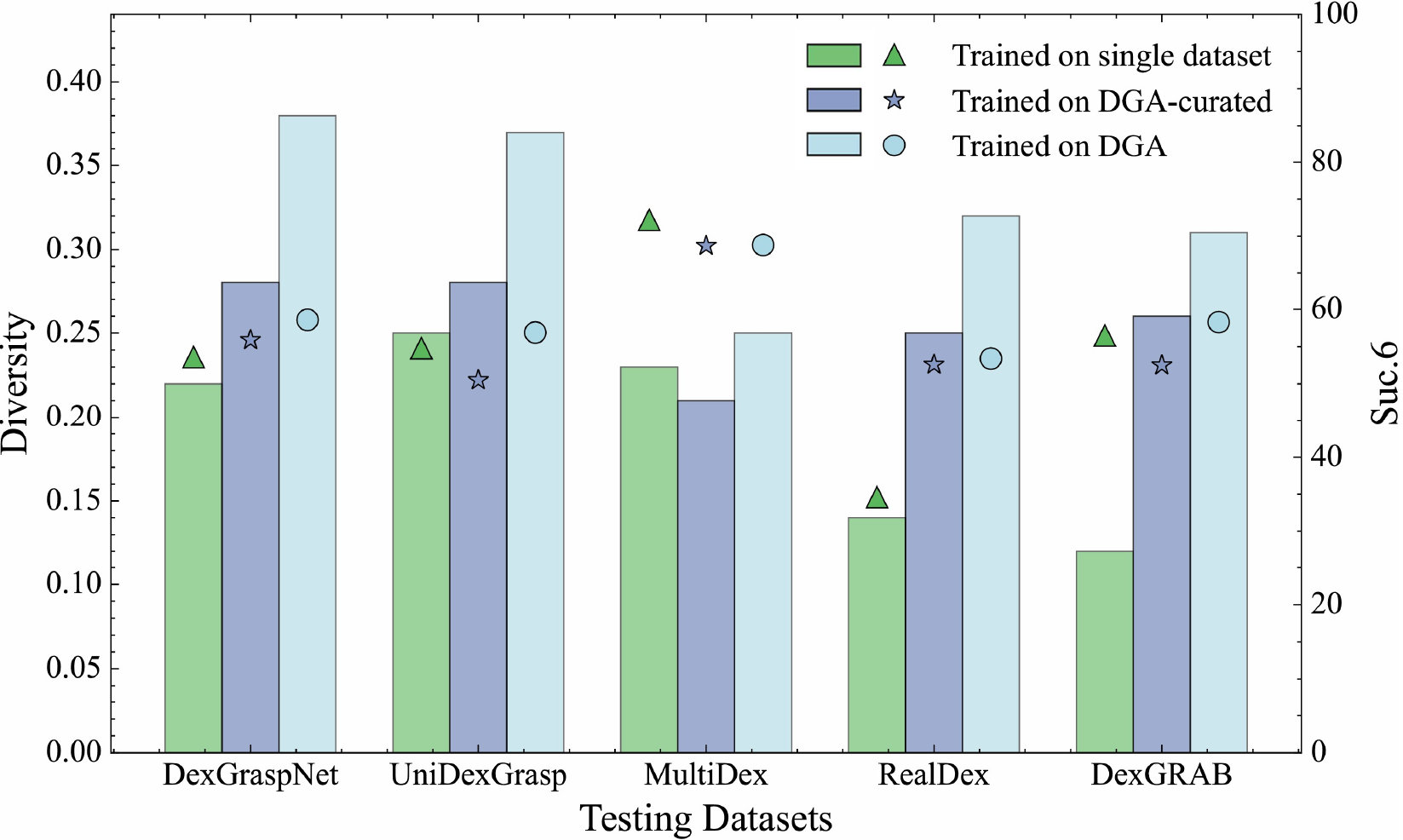}
    \caption{Cross-dataset evaluation. Comparison of diversity (bars) and all-direction grasp success rates (triangles/stars/circles) across models trained on different datasets. Trained on single dataset indicates models were trained on the same dataset they are tested on.}
   \label{fig:single-curated-dga}
\end{figure}
\section{Evaluation Results}\label{sec: sup-result}
\subsection{Results for Cross-dataset Evaluation}
\begin{figure*}[ht]
  \centering
   \includegraphics[width=1.0\linewidth]{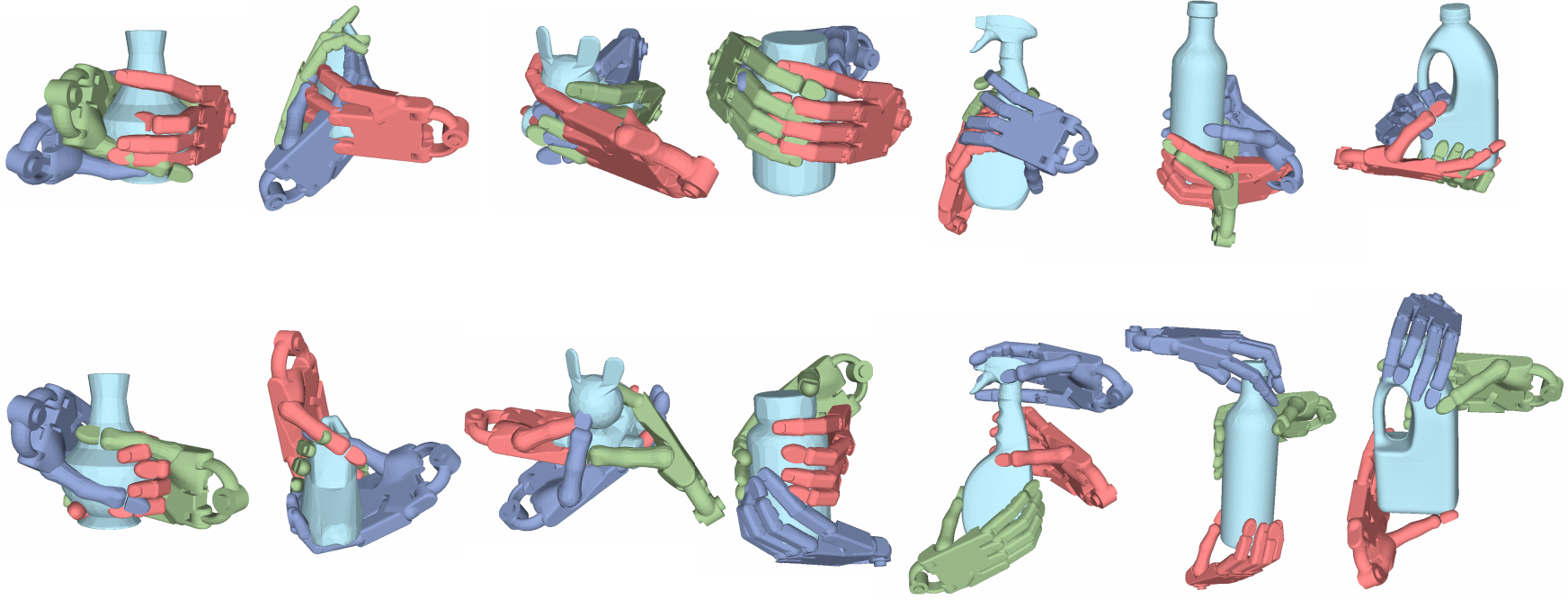}
    \caption{Visualization of cross-dataset evaluation results. The top row shows models trained on single dataset,while the bottom row displays models trained on our dataset.}
   \label{fig:visual_cross_datasets}
\end{figure*}

We present the comprehensive cross-dataset evaluation result for the DexGrasp Anything diffusion generator on five existing datasets and our dataset, as shown in Table~\ref{tab:cross-dataset} and Figure~\ref{fig:single-curated-dga}. Qualitative results are presented in Figure~\ref{fig:visual_cross_datasets}. The results demonstrate that training on our large-scale, diverse dataset significantly enhances generation diversity while achieving comparable or higher grasping success rates compared to training on the respective original datasets.

\begin{figure}[htbp]
  \centering
   \includegraphics[width=1.0\linewidth]{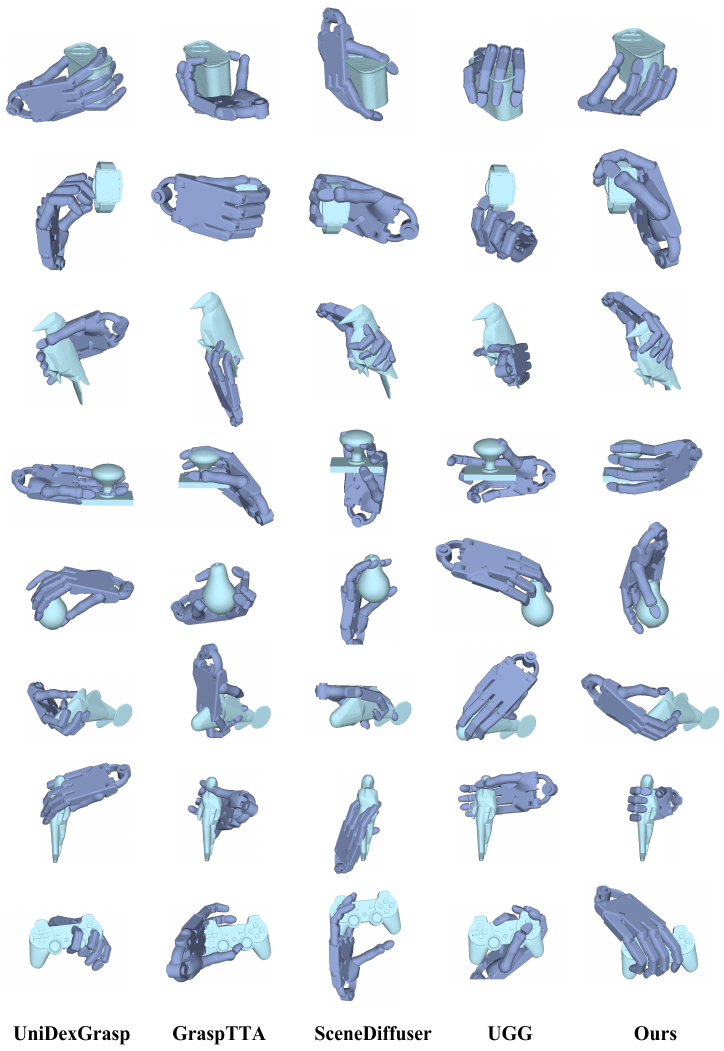}

    \caption{Qualitative visualization of comparisons on grasping poses.}
   \label{fig:sup_table2}
\end{figure}

\subsection{Qualitative Results for Comparisons}
We provide additional qualitative results comparing DexGrasp Anything with existing state-of-the-art methods in Figure~\ref{fig:sup_table2}. 

\begin{figure}[htbp]
  \centering
   \includegraphics[width=1.0\linewidth]{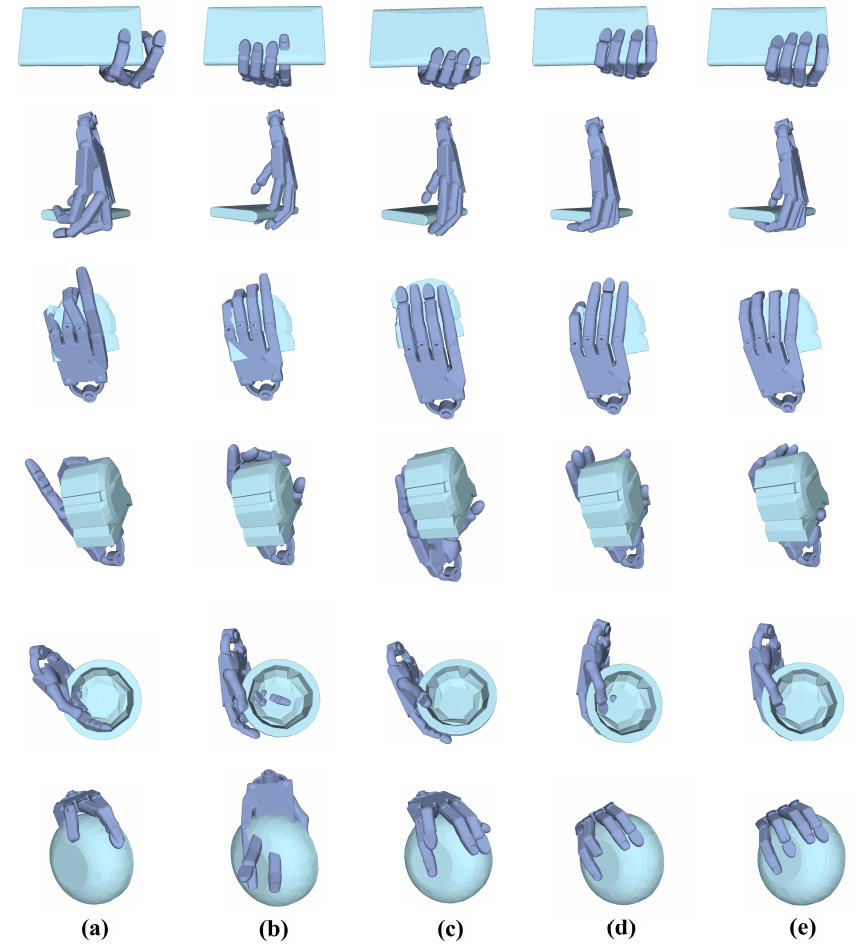}

    \caption{Visualizations of the ablation study. Each pair(row 1-2, 3-4, 5-6) of rows corresponds to the different views of the same grasp for the same object.}
   \label{fig:sup_ablation}
\end{figure}

\subsection{More Visualizations for Ablation Studies}
We provide additional visualizations for the ablation studies in Figure~\ref{fig:sup_ablation}, where we progressively incorporate three physical constraints during the training and sampling process of our diffusion generator. Column (a) represents the baseline, while columns (b), (c), and (d) illustrate the results after incrementally adding the SRF, ERF, and SPF constraints, respectively, to the baseline. Finally, column (e) shows the results after incorporating the LLM-enhancement into the representation extraction module.

\begin{table*}[htbp]
\caption{Cross-dataset evaluation results. The \textbf{bold} values indicate the best performance, and the \underline{underlined} values indicate the second-best performance.}
\centering
\resizebox{\linewidth}{!}{
\begin{tabular}{@{}l@{\hspace{0.2cm}}*{20}{@{\hspace{0.1cm}}c}@{}}
\toprule
\textbf{Testing Dataset} & \multicolumn{4}{c}{DexGraspNet} & \multicolumn{4}{c}{UniDexGrasp} & \multicolumn{4}{c}{MultiDex} & \multicolumn{4}{c}{RealDex} & \multicolumn{4}{c}{DexGRAB}\\ 
\cmidrule(lr){2-5} \cmidrule(lr){6-9}\cmidrule(lr){10-13}\cmidrule(lr){14-17}\cmidrule(lr){18-21}
\textbf{Training Dataset} &\textbf{Suc.6 $\uparrow$} & \textbf{Suc.1 $\uparrow$} & \textbf{Pen. $\downarrow$} & \textbf{Div $\uparrow$} &\textbf{Suc.6 $\uparrow$} & \textbf{Suc.1 $\uparrow$} & \textbf{Pen. $\downarrow$} & \textbf{Div $\uparrow$}& \textbf{Suc.6 $\uparrow$} & \textbf{Suc.1 $\uparrow$} & \textbf{Pen. $\downarrow$} & \textbf{Div $\uparrow$}& \textbf{Suc.6 $\uparrow$} & \textbf{Suc.1 $\uparrow$} & \textbf{Pen. $\downarrow$} & \textbf{Div $\uparrow$}& \textbf{Suc.6 $\uparrow$} & \textbf{Suc.1 $\uparrow$} & \textbf{Pen. $\downarrow$} & \textbf{Div $\uparrow$}  \\ 
\midrule
DexGraspNet & 53.6 & \textbf{90.4} & 21.5 & 0.22 & 49.3 & 82.4 & 14.9& 0.19 & 55.6 & 90.1 & \textbf{9.1} & 0.17 & 38.4 & 77.5 & \textbf{19.2} & 0.17 & 48.1 & 84.0 & \textbf{19.7} & 0.18  \\ 
UniDexGrasp & 45.4 & 82.4 & \textbf{16.4} & 0.23 & \underline{54.8} & \textbf{90.8} & \underline{18.9} & 0.25 & 52.8 & 90.3 & 9.4 & 0.18 & 38.4 & 79.3 &  20.7 & 0.19 & 37.5 & 79.6 & \underline{20.1} & 0.19  \\ 
MultiDex & 46.8 & 83.1 & 18.1 & 0.20 & 43.9 & 81.3 & 14.5 & 0.19 & \textbf{72.2} & \underline{96.3} & 9.6 & \underline{0.23} & 29.6 & 69.1 & \underline{20.1} & 0.23 & 52.9 & 87.9 & 21.0 & 0.15  \\ 
RealDex & 47.3 & 79.5 & 18.7 & 0.05 & 43.8 & 81.3 & 15.8 &  0.04 & 57.5 & 89.2 & 11.6 & 0.06 & 34.6 & 71.2 & 23.1 & 0.14 & 38.5 & 79.8 & 22.7 & 0.08   \\ 
DexGRAB & 41.0 & 75.8 & 18.7 & 0.12 & 43.9 & 81.4 & 14.1  & 0.10 & 62.1 & 90.9 & \underline{9.3} & 0.11 & 35.2 & 71.5 & 24.0 & 0.11 & \underline{56.5} & \textbf{91.8} & 28.6 & 0.12   \\ 
\midrule 
DGA-curated(ours)& \underline{55.9} & 87.3 & 20.9 & \underline{0.28} & 50.5 & 84.6 & \textbf{14.0} & \underline{0.28} & 68.7 & 95.9 & 12.5 & 0.21 & \underline{52.6} & \textbf{85.7} & 21.5 & \underline{0.25} & 52.5 & 89.0 & 22.9 & \underline{0.26} \\
DGA(ours) & \textbf{58.6} & \underline{88.5} & \underline{17.8} & \textbf{0.38} & \textbf{56.9} & \underline{86.7} & 16.7 & \textbf{0.37} & \underline{68.8} & \textbf{96.9} & 9.5 & \textbf{0.25} & \textbf{53.4} & \underline{84.4} & 22.4 & \textbf{0.32} & \textbf{58.3} & \underline{90.2} & 23.2 & \textbf{0.31} \\
\bottomrule
\end{tabular}}
\label{tab:cross-dataset}
\end{table*}

\subsection{More Visualizations of Generated Poses}
\begin{figure*}[ht]
  \centering
   \includegraphics[width=1.0\linewidth]{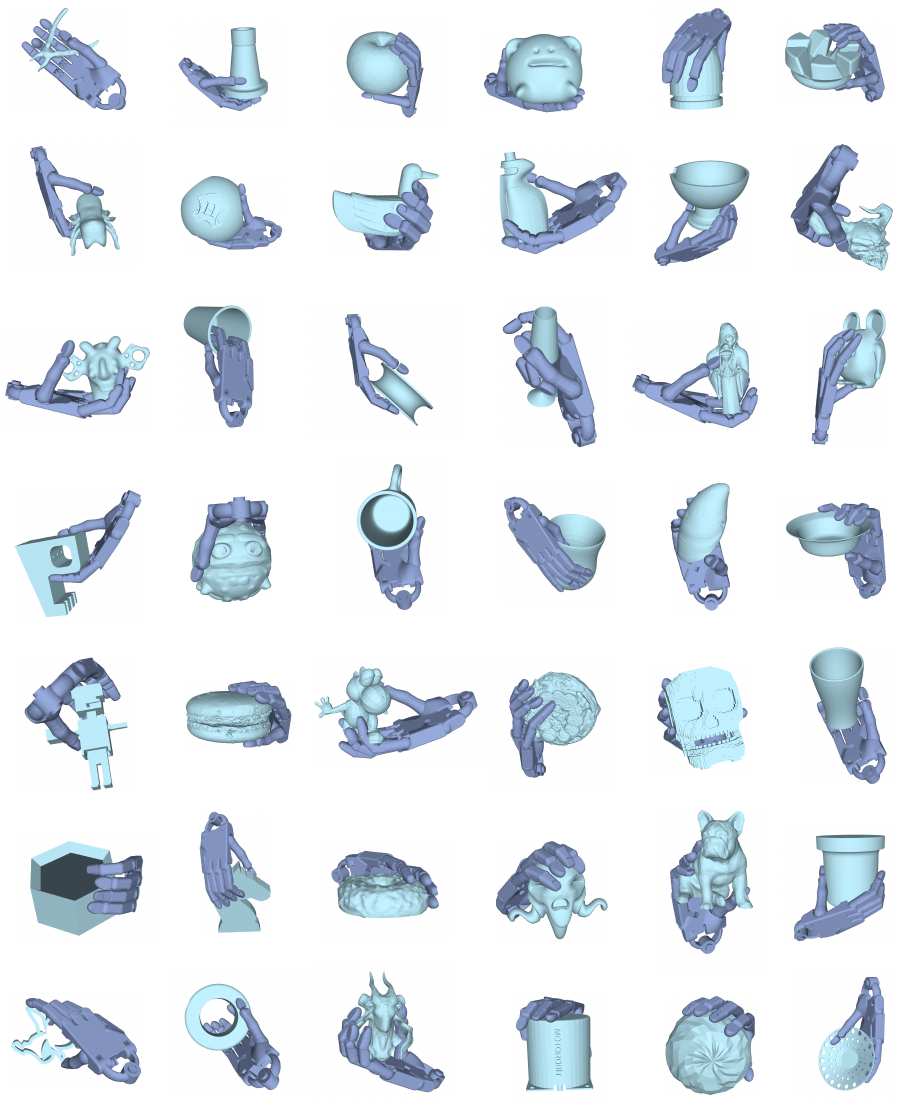}
    \caption{Visualization of  our method’s results.}
   \label{fig:visual_our_dataset}
\end{figure*}

\begin{figure*}[ht]
  \centering
   \includegraphics[width=1.0\linewidth]{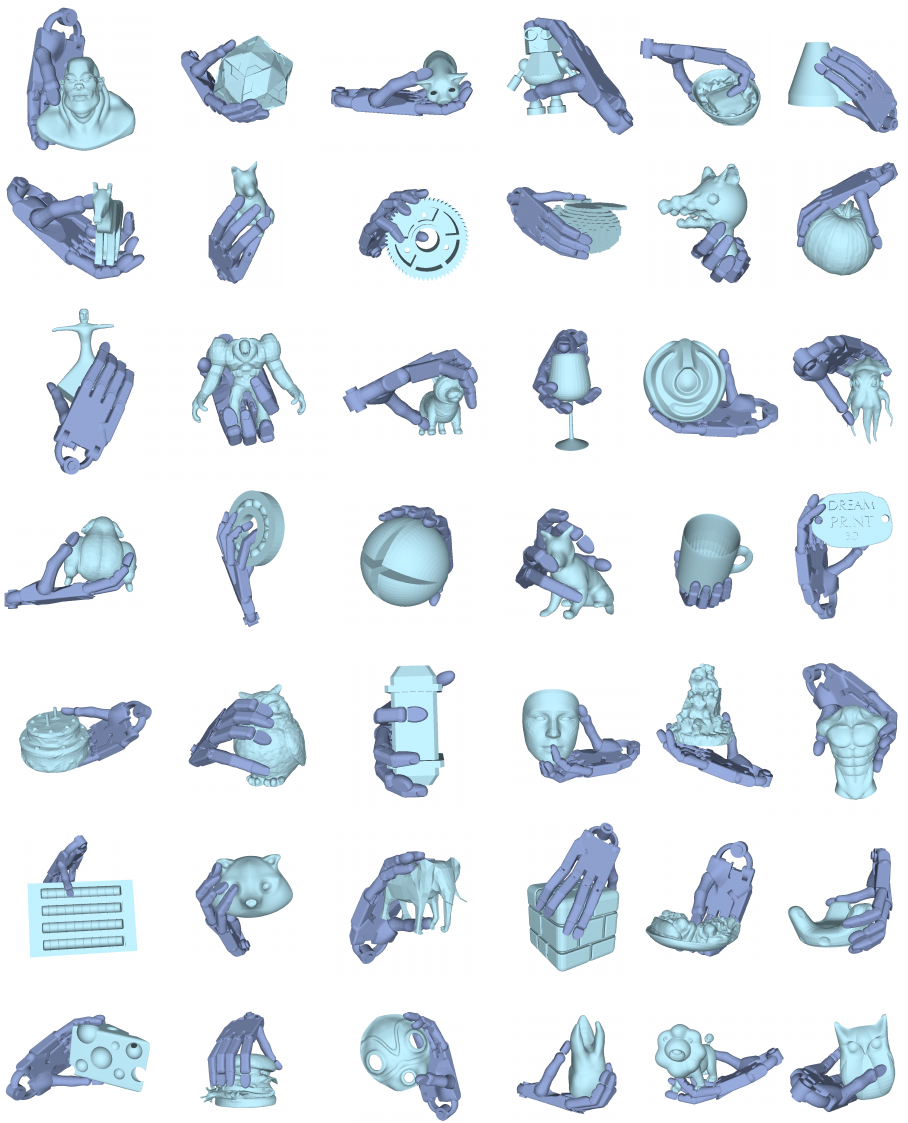}
    \caption{Visualization of  our method’s results.}
   \label{fig:visual_our_dataset_2}
\end{figure*}

We present more visualizations of the generated grasping poses by our methods on various challenging objects from~\cite{deitke2023objaverse, deitke2024objaverse} in Figure~\ref{fig:visual_our_dataset} and Figure~\ref{fig:visual_our_dataset_2}. Our models produce reasonable and stable grasping poses for complex and irregular objects such as a robot model (3rd row, 2nd col in Figure~\ref{fig:visual_our_dataset_2}) and a loong head (7th row, 3rd col in Figure~\ref{fig:visual_our_dataset_2}).

\begin{figure}[ht]
  \centering
   \includegraphics[width=1.0\linewidth]{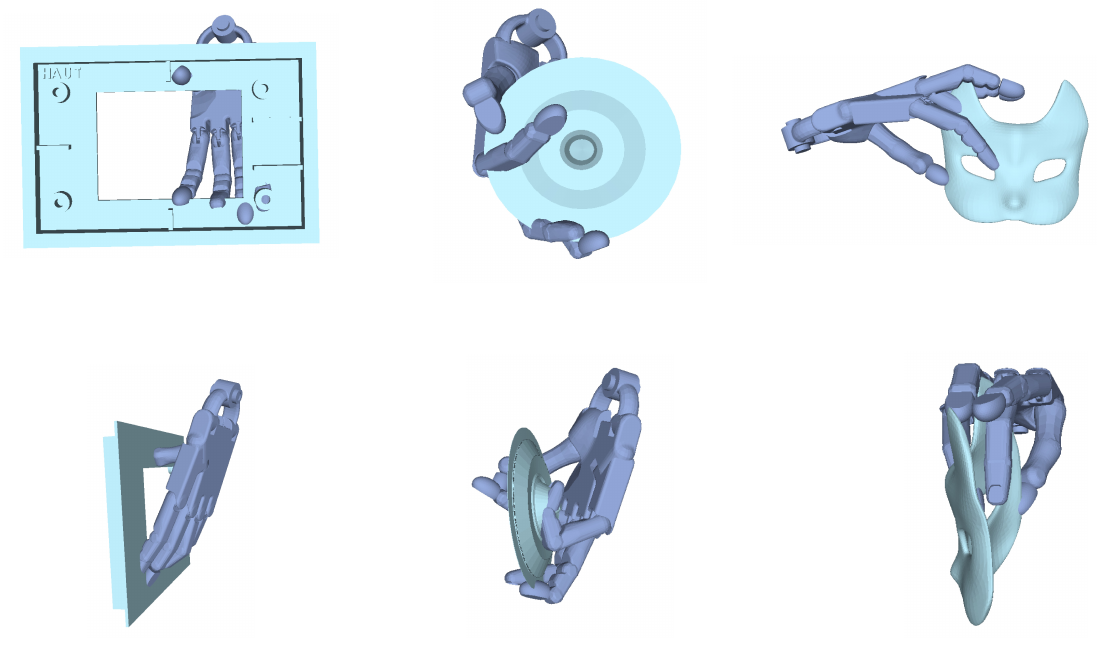}
    \caption{Visualization of failed cases.}
   \label{fig:failed results}
\end{figure}

\section{Implementation details}\label{sec: details}
We rigorously evaluated each grasp pose in our dataset to ensure that the object is held firmly without significant penetration. For hand-object penetration computation, we employ two approaches. The first approach, adopted by \cite{scenediffuser} and also used in the External-penetration Repulsion Force, calculates the Euclidean distance between each hand point and its nearest neighbor on the object surface. The second approach, introduced by \cite{xu2023unidexgrasp}, transforms the object and each robot hand link into the local hand coordinate system based on the robot's configuration. For the palm, penetration is measured as the signed distance between the sampled object points and the mesh surface of the palm, represented by a signed distance field. For each phalange link, it is approximated as cylinders, and object points are projected onto the cylinders' bounding volumes to compute signed distances, adjusted using a mask to differentiate internal and external points. We combine both methods to enforce strict filtering conditions for our dataset.

\section{Limitations and Future Works}
As shown in Figure~\ref{fig:failed results}, we notice that our method produces sub-optimal poses with obvious penetration for objects with extremely thin shapes(e.g. masks, plates etc.). To address these challenges, enhancing affordance modeling or integrating tactile feedback into the robotic grasping system would be promising directions for future works.


\end{document}